\documentclass[journal]{IEEEtran}
\usepackage[hidelinks]{hyperref}
\usepackage{amsthm}
\usepackage{amssymb}
\usepackage{mathtools}
\usepackage{mathrsfs}
\usepackage[justification=centering]{caption}
\usepackage{caption}
\usepackage{booktabs}
\usepackage{multirow}
\usepackage[table]{xcolor}
\usepackage[ruled,linesnumbered]{algorithm2e}
\usepackage{subfigure}
\usepackage{float}
\usepackage{bbm}
\usepackage{hyperref}

\begin{document}
    
    \title{Model Predictive Control for Trajectory Tracking  on Differentiable Manifolds}
    
    \author{Guozheng Lu, Wei Xu, Fu Zhang
        \thanks{This project is supported by Hong Kong RGC ECS under grant 27202219. 
        
        All authors are with Mechatronics and Robotic Systems (MaRS) Laboratory, Department of Mechanical Engineering, University of Hong Kong. {\tt\small \{gzlu, xuweii, fuzhang\}@hku.hk}.}
    }

    \maketitle
    
    \begin{abstract}

        We consider the problem of bridging the gap between geometric tracking control theory and implementation of model predictive control (MPC) for robotic systems operating on manifolds.  We propose a generic on-manifold MPC formulation based on a canonical representation of the system evolving on manifolds. Then, we present a method that solves the on-manifold MPC formulation by linearizing the system along the trajectory under tracking. There are two main advantages of the proposed scheme. The first is that the linearized system leads to an equivalent error system represented by a set of minimal parameters without any singularity. Secondly, the process of system modeling, error-system derivation, linearization and control has the manifold constraints completely decoupled from the system descriptions, enabling the development of a symbolic MPC framework that naturally encapsulates the manifold constraints. In this framework, users need only to supply system-specific descriptions without dealing with the manifold constraints. We implement this framework 
        and test it on a quadrotor unmanned aerial vehicle (UAV) operating on $SO(3) \times \mathbb{R}^n$ and an unmanned ground vehicle (UGV) moving on a curved surface. Real-world experiments show that the proposed framework and implementation achieve high tracking performance and computational efficiency even in highly aggressive aerobatic quadrotor maneuvers.
    \end{abstract}


    \IEEEpeerreviewmaketitle

    \section{Introduction}
    %
    %
    %
    %
    %
    
    \IEEEPARstart{T}{he} configuration space of robotic systems usually involve manifolds. For instance, $\textbf{1)}$ Lie group: the rotation workspace of satellites \cite{crouch1984spacecraft} and three-axis camera stabilizers \cite{lu2019imu} is lying on the special orthogonal group $SO(3)$; the rigid rotation and translation workspace of manipulators \cite{murray2017mathematical} and multi-coptor UAVs \cite{mahony2012multirotor} is the special Euclidean group $SE(3)$; the movement of cars on a plane is considered as $SE(2)$. $\textbf{2)}$ 2-sphere $\mathbb{S}^2$: 3D pendulum movement of pointing devices \cite{bullo1999tracking}; $\textbf{3)}$ 2D surface: cars are always restricted to move on ground surfaces, which have only two degree of freedom \cite{zhang2021pose}; $\textbf{4)}$ robot swarms that include the above geometry structures.
    
    On the other hand, MPC as an advanced control technique for multivariable systems with state and input constraints, has recently attained fruitful success in robotic applications, such as autonomous driving \cite{falcone2007linear}, \cite{williams2018information}, UAVs \cite{kamel2017linear}, \cite{bangura2014real}, mobile manipulators \cite{minniti2019whole}, legged robots \cite{neunert2018whole}, \cite{di2018dynamic} and underwater robots \cite{fernandez2016model}. MPC is an inclusive framework for defining objective functions and constraints, making it amenable to complex autonomy, like learning \cite{bouffard2012learning}, \cite{zhang2016learning}, obstacle avoiding \cite{lindqvist2020nonlinear}, \cite{liu2017combined}, perception awareness \cite{falanga2018pampc} and swarm collaboration \cite{luis2020online}. However, MPC is usually designed for systems whose state is described in flat vector $\mathbb{R}^n$. When deploying it to robotic systems evolving on curved, non-vector-space manifolds, specific treatment has to be taken. For example, to cope with $SO(3)$, existing works usually use minimal-parametrization (e.g., Euler angles \cite{kamel2017linear}) that suffers from singularity, or over-parametrization (e.g., quaternion \cite{falanga2018pampc} or rotation matrix \cite{kalabic2017mpc}) that brings in extra normalization constraints to the state. These trivial treatments need to be designed carefully for a certain system (or manifold) and usually cannot be generalized to different manifolds. Moreover, theories and experiences developed for MPC formed in $\mathbb{R}^n$ cannot  be immediately extended to manifolds. To the best of our knowledge, there are very few studies that formally consider the design of MPC on generic manifolds. 
    
    \begin{figure}[!t] 
        \centering
        \includegraphics[width=1\linewidth]{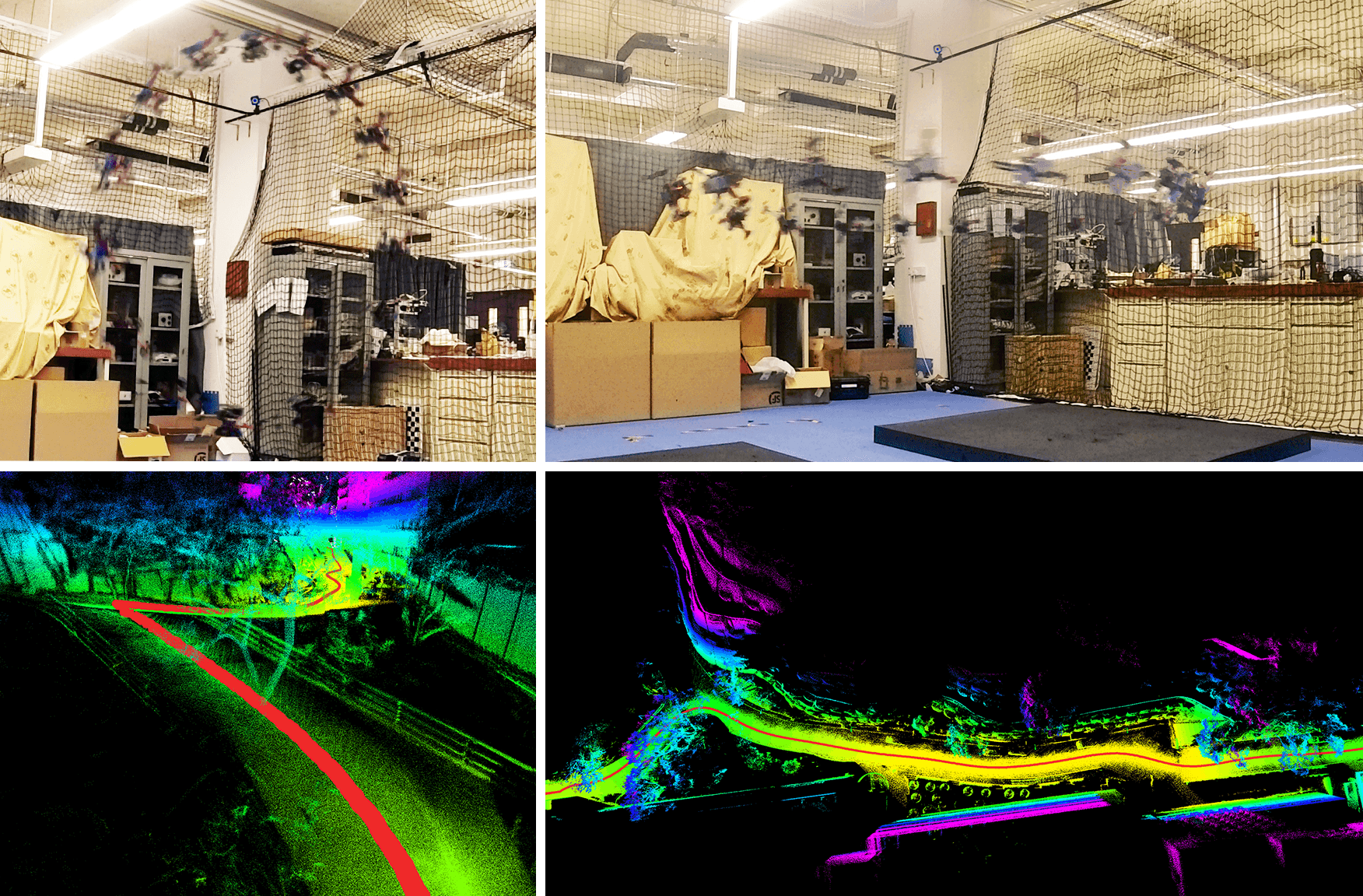} 
        \caption{Example applications of our MPC to robotic systems evolving on different manifolds. The above is aggressive UAV aerobatics, and the below is curved UGV trajectories on the lidar map.  Video: \url{https://youtu.be/wikgrQbE6Cs}}
        \label{f:quad_car_traj}
        \setlength{\abovecaptionskip}{-0.1cm} 
        \setlength{\belowcaptionskip}{-0.2cm}
        \vspace{-0.3cm}
    \end{figure}
    
    In this study, we show that it is possible to integrate the manifold constraint into the system modeling and MPC formulation. An overview of our  contributions are as follows:
    \begin{itemize}
        \item [1)] A formal and generic MPC formulation for trajectory tracking control of systems evolving on manifolds.
        
        \item [2)] A novel method that solves the on-manifold MPC formulation. The method completely decouples the manifold constraints from the system descriptions, enabling the development of a symbolic MPC framework that naturally encapsulates the manifold constraints. 
        
        \item [3)] Evaluation of the proposed formulation and implementation on real-world systems evolving on different manifolds. Experiments demonstrate the superior efficiency and high tracking performance even in challenging UAV aerobatics (see Fig. \ref{f:quad_car_traj}).
        
    \end{itemize}

    The rest of the paper is organized as follows. Section \ref{sec:related_works} gives a review of related works to the posed problem. In Section \ref{sec:operations}, brief differential geometry and the proposed operations of $\boxplus/\boxminus$ and $\oplus$ are introduced. Section \ref{sec:canonical_representation} proposes a formal, generic and on-manifold MPC formulation for trajectory tracking. Section \ref{sec:mpc} details the efficient solution of the on-manifold MPC. In Section \ref{sec:experiment}, experimental results on two real-time robotic systems demonstrate the tracking performance and efficiency of the proposed framework. Finally, conclusion and future work are arrived at Section \ref{sec:conclusion}.

    \section{Related Works}  
    \label{sec:related_works}
    Related works can be selected along three four directions: nonlinear control, geometric tracking control, geometric linearization methods, and MPC for trajectory tracking applied to robotic systems.
    
    \subsection{Nonlinear control}
    
    A naive way to eliminate the manifold constraints is to parameterize a manifold by minimal parameters. The resultant system model based on the minimal parametrization is generally nonlinear and can be controlled using any nonlinear control approaches. While being straightforward, this method completely gives up the intrinsic geometry structure of manifolds. Moreover, due to the topological obstruction, a minimal parametrization always suffers from certain singularities and significantly restricts the system operation range. For example, the rotational special orthogonal group $SO(3)$ can be minimally represented by three Euler angles, but would encounter singularities when the second angle is at $\pm 90^\circ$. 
    
    \subsection{Geometric Control}
    Beginning in the 1970s, the aerospace problems at the time asked a new theoretical standpoint that can unveil both geometry and mechanics to the early nonlinear control theory and led to the emergence of geometric control  \cite{brockett2014early}. In particular, trajectory tracking posed a challenge that how to steer the dynamic system along reference trajectories with guaranteed convergence while respect the manifold constraints. Bullo and Muarry \cite{bullo1999tracking} proposed a generic geometric tracking control framework for fully actuated mechanical systems on manifolds. This work presented an intrinsic geometric description of compatible configuration and velocity error on manifolds by the definition of an error function and a transport map. A proportional derivative feedback and feedforward controller for trajectory tracking was proven with almost global stability and local exponential convergence. Specifically, for attitude tracking on $SO(3)$, Mayhew {\it et al} \cite{mayhew2009robust} first proposed a quaternion-based control.  To avoid the unwinding phenomenon \cite{bhat2000topological}, Lee \cite{lee2011geometric} further developed a geometric control theory on $SO(3)$ that the tracking controller guarantees the exponential stability even involving a large initial attitude error.  Benefiting from the promising theoretical results, extensive successful real-word experiments were presented, including the tracking control for quadrotor on $SE(3)$ \cite{lee2010geometric} and quadrotor with a cable-suspended load on $SE(3) \times \mathbb{S}^2$ \cite{sreenath2013geometric}. While these works have established the foundation of geometric control theory, they almost all focused on the geometric structure of manifolds, and only used very simple controllers (e.g., proportional–integral–derivative (PID) or its variants) to ease the stability analysis.

    \subsection{Geometric Linearization Methods}
    To further extend the concept of geometric control to more advanced control techniques like linear quadratic regulator (LQR), geometric linearization methods haven been studied in the area of control theory. For instance, a geometric Jacobian linearization framework and LQR theory for general manifold was presented in \cite{lewis2010geometric}, and the optimal control on Lie group was studied in \cite{saccon2013optimal}. However, these previous works were abstract and focused on the time-invariant linearization about a static equilibrium, thus introducing not many insights into robotic systems tracking time-varying trajectories. More recently, a variation-based linearization method developed on \cite{lee2008computational} was introduced by Wu {\it et al} \cite{wu2015variation} to formulate the variation dynamics based on the original system evolving on $SO(3)$ and $\mathbb{S}^2$, which is valid globally and singularity-free for tracking time-varying trajectories. 
    
    Compared with the variation-based method above, our proposed error-state linearization scheme is somewhat equivalent, but forms the system and control problem all in discrete time (as opposed to continuous time in \cite{lewis2010geometric, lee2008computational, wu2015variation}) that naturally supports more advanced model predictive control (MPC). Moreover, our formulation is derived based on basic calculus, making it more intuitive for engineering practitioners with few preliminaries about differential geometry. This geometric approach is inspired by error-state extend Kalman filter (ESEKF \cite{lu2019imu}, or named multiplicative EKF \cite{markley2003attitude} , indirect EKF\cite{trawny2005indirect}), and actually not the first time to be applied to robotic control systems, e.g. Hong \cite{hong2020real} has verified the nonlinear MPC on $SO(3)$ for legged locomotion in a simulation environment.
    
    \subsection{MPC on Robotic Systems}
    MPC has enjoyed significant success and widespread use in process industry for decades, and recently brings the benefits into robotic applications. MPC can simultaneously emphasizes tracking performance and satisfies input and state constraints by solving a finite horizon optimal control problem online and applying the first element of the optimized input sequence to the system \cite{borrelli2017predictive}. To apply MPC to robotic systems evolving on manifolds, researchers came up with usable but not complete solutions. For example, Nguyen {\it et al} \cite{Nguyen2020model} parameterized the attitude by three Euler angles which suffer from the singularity issue mentioned above. To avoid the singularity, Falanga {\it et al} \cite{falanga2018pampc} uses a quaternion parametrization, but needs further to add one extra constraints (i.e., unit length) to the state in the MPC. Kalabić \cite{kalabic2017mpc} \cite{kalabic2016mpc} use a full rotation matrix to represent an attitude in the cost function of MPC, but the cost weight tuning is less intuitive due to the over-parametrization in rotation matrix. Compared with these works, our method is applicable to general manifolds and suffer from no singularity or over-parametrization problems.  
    
    Another challenge of implementing a MPC on real-time systems is the substantial computation load by solving the optimization problem online. Optimization problems for tracking control in existing literature are usually expressed as a Nonlinear Programming (NLP) that can be solved by Sequential Quadratic Programming (SQP) to make a compromise between optimality and efficiency. In some lightweight implementation such as UAVs, technical tricks like increasing the sample time in the discrete time model to ensure a high-rate control \cite{kamel2017linear}, are taken to further reduce the computational time. Although constraints are satisfied, the demonstrated tracking performance of MPC \cite{kamel2017linear},\cite{foehn2018onboard} did not show the comparative tracking accuracy with conventional linear controllers (e.g. PID controller) \cite{mellinger2011minimum}, \cite{faessler2017differential}.

    \section{Operations on Differentiable Manifolds}\label{sec:operations}
    
    In this section we provide mathematical background of basic differential geometry and operations needed for the remainder of the paper. We refer to \cite{boothby1986introduction}, \cite{robbin2011introduction} for a mathematical introduction to differential geometry, and \cite{he2021embedding} for more details for operators of $\boxplus / \boxminus$ and $\oplus$. Examples of important manifolds are also provided.
    
    \subsection{Differentiable Manifolds}
    A topological $n$-manifold is a topological space $\mathcal{M}$ with dimension of $n$ such that each point $p \in \mathcal{M}$ has an open neighborhood $U$ that is homeomorphic to an open subset of Euclidean space $\mathbb{R}^n$ (the {\it homeomorphic space}). The manifold can be parameterized by a set of local coordinate charts $(U, \varphi)$ where $U \subset \mathcal{M}$, $\varphi$ is a map projecting the manifold to the linear homeomorphic space $\mathbb{R}^n$ that $\varphi: \mathcal{M} \mapsto \mathbb{R}^n$ and inversely $\varphi^{-1}$: $\mathbb{R}^n \mapsto \mathcal{M}$. If any two overlapping charts $(U, \varphi)$ and $(V, \psi)$ are $C^k$-compatible\footnote{$(U, \varphi)$ and $(V, \psi)$ are $C^k$-compatible if $U \cap V$ is nonempty and the composite map $\phi \circ \psi^{-1}$ is $C^k$ smooth.} \cite{boothby1986introduction}, the topological manifold is a $C^k$-differentiable manifold. A vector space that consists of all directional derivatives across point $p$ is called the tangent space locally at $p$, denoted $T_p\mathcal{M}$.
    
    \begin{figure}[!t] 
        \centering
        \includegraphics[width=0.6\linewidth]{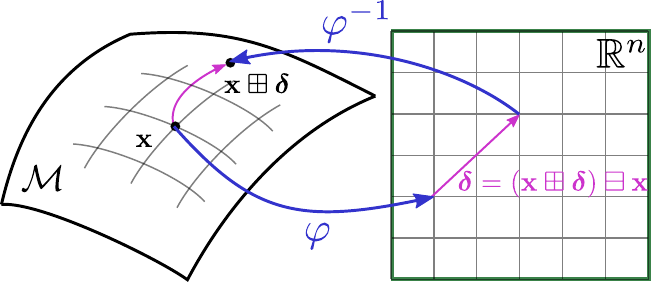} 
        \caption{Illustration of the $\boxplus/\boxminus$ operations on manifold with a perturbation in its homeomorphic space.}
        \label{f:manifold_operation}
        \setlength{\abovecaptionskip}{-0.0cm} 
        \setlength{\belowcaptionskip}{-0.2cm}
        \vspace{-0.3cm}
    \end{figure}
    
    \subsection{Operations of $\boxplus / \boxminus$}
    Based on the property of manifold that it is locally homeomorphic to $\mathbb{R}^n$, \cite{hertzberg2013integrating} established a generic bi-directional mapping between a manifold in the neighborhood centering at a given point on $\mathcal{M}$ and a small, coordinate-free perturbation in $\mathbb{R}^n$. Two encapsulated operators $\boxplus / \boxminus$ are implemented to this mapping:
    \begin{eqnarray}
        \boxplus : \mathcal{M} \times \mathbb{R}^n \mapsto \mathcal{M} ,\qquad 
        \boxminus : \mathcal{M} \times \mathcal{M} \mapsto \mathbb{R}^n 
    \end{eqnarray}
    
    Consider $\mathbf{x}$ be a point on $\mathcal{M}$ and $\mathbf{y}$ locates in the neighborhood $U \subset \mathcal{M}$ of $\mathbf{x}$. $\boldsymbol{\delta}$ is a small, coordinate-free perturbation in $W \subset \mathbb{R}^n$ that is homeomorphic to $U$. As is visualized in Fig. \ref{f:manifold_operation}, the following axioms are held for every $\mathbf{x} \in \mathcal{M}$:
    \begin{eqnarray}
        \mathbf{x} \boxplus \mathbf{0} = \mathbf{x} ,\quad
        \mathbf{x} \boxplus (\mathbf{y} \boxminus \mathbf{x}) = \mathbf{y} , \quad
        (\mathbf{x} \boxplus \boldsymbol{\delta}) \boxminus \mathbf{x}  = \boldsymbol{\delta}
        \label{e:axiom}
    \end{eqnarray}
    
    The homeomorphic space $\mathbb{R}^n$ representing perturbations in $\boxplus\backslash\boxminus$ can be chosen as any local coordinate spaces and in particular the minimal parametrization of the tangent space $T_\mathbf{x}\mathcal{M}$. The union of homeomorphic space parametrization at each point on the manifold is globally free of singularity because the neighbourhood at each point on a differentiable manifold is always compatible, while a single local coordinate may exhibit singularities. For instance, on 2-sphere $\mathbb{S}^2$, the latitude-longitude parametrization can be considered as one  local coordinate and is singular at the pole.
    
    \subsection{Operation of $\oplus$}
    Different from defining the operation $\boxplus$ to drive the manifold configuration by a perturbation in the homeomorphic space, in practical systems, the perturbation on state caused by exogenous input is not always inherently lying in the defined homeomorphic space. An example on robotic systems is that the gravity vector \cite{he2021embedding} considered as on the 2-sphere $\mathbb{S}^2$, is subject to a 3-dimension angular velocity which is independent of the sphere tangent space that defines $\boxplus$. In order to describe the configuration change due to an  exogenous perturbation $\boldsymbol{\delta}^e$, we introduce a operation $\oplus$:
    \begin{equation}
        \small
        \oplus : \mathcal{M} \times \mathbb{R}^l \mapsto \mathcal{M}
        \setlength{\abovedisplayskip}{0.15cm} 
        \setlength{\belowdisplayskip}{0.15cm} 
    \end{equation}
    where it is not necessarily $l = n$. If the exogenous perturbation does lie in the homeomorphic space of $\boxplus$, the two operations are equivalent: $\oplus \simeq \boxplus: \mathcal{M} \times \mathbb{R}^n \mapsto \mathcal{M}$.
    
    \subsection{Examples of Practically Important Manifolds}
    With the above definitions, $\boxplus$ gives a description that a manifold configuration arrives at another steered by a small perturbation in the homeomorphic space $\mathbb{R}^n$. Conversely, the difference between two manifold configurations can be mapped into the homeomorphic space $\mathbb{R}^n$ by $\boxminus$. In practise, if there is exogenous perturbation beyond the homeomorphic space defined by $\boxplus$, $\oplus$ is introduced to denote the consequently shifted configuration. Follows are examples of the operations $\boxplus / \boxminus$ and $\oplus$ on several practically important manifolds.
    
    \subsubsection{\textbf{\textit{Example 1: Euclidean Space $\mathbb{R}^n$}}}~
    
    For $\mathcal{M} = \mathbb{R}^n$, the tangent space is itself at any point and based on which, the operation of $\boxplus / \boxminus$ are simple addition and subtraction.
    \begin{eqnarray}
        \mathbf{x} \boxplus \boldsymbol{\delta} = \mathbf{x} + \boldsymbol{\delta} ,\qquad
        \mathbf{y} \boxminus \mathbf{x} = \mathbf{y} - \mathbf{x}     
    \end{eqnarray}
    The exogenous perturbation lies also in $\mathbb{R}^n$, the tangent space defining $\boxplus$, making $\oplus / \boxplus$ the same.

    \subsubsection{\textbf{\textit{Example 2: Lie Groups $\mathit{G}$}}} ~
    
    If $\mathcal{M}$ is a matrix Lie group (denoted as $G$) embedded in $\mathbb{R}^{n \times n}$, the tangent space at a configuration $g \in \mathit{G}$ is $T_{g}\mathit{G} := g \xi$ for $\xi \in  \mathfrak{g}$, where $\mathfrak{g}$ is the Lie algebra of $G$ \cite{robbin2011introduction}. Define  $\text{Exp}(\cdot)$ the map from $\mathbb{R}^n$ parameterizing $\mathfrak{g}$ to $G$ \cite{murray2009optimization} and its inverse $\text{Log}(\cdot)$, the $\boxplus\backslash\boxminus$ are then:
    
    \begin{eqnarray}
        &\mathbf{x} \boxplus \boldsymbol{\delta} = \mathbf{x} \cdot  \text{Exp}(\boldsymbol{\delta}) , \qquad \mathbf{y} \boxminus \mathbf{x} = \text{Log}\left(\mathbf{x}^{-1}\mathbf{y}\right)    
        \label{e:lie_group}
    \end{eqnarray}
    
    
    In practical system evolving on Lie groups, exogenous perturbation (e.g., angular rate for $SO(3)$ and twist for $SE(3)$ over an infinitesimal time period) naturally lies in the tangent space defining $\boxplus$. Thus, we have $\oplus \simeq \boxplus$.
    
    Specifically, Lie groups of special orthogonal group $SO(n) \triangleq \{\mathbf{R}\in \mathbb{R}^{n\times n} | \mathbf{R}^T\mathbf{R} = \mathbf{I}, \det\mathbf{R} = 1\}$ and special Euclidean group $SE(n)$ for $n = 2, 3$ are the most common work spaces where robotic systems operating with rotation and translation. Since $SE(n)$ can be viewed as a compound manifold $SE(n) := \mathbb{R}^n \times SO(n)$, we just focus on $SO(2)$ and $SO(3)$ for the sake of simplicity.
    \begin{subequations}
        \small
        \begin{align}
            &\text{Exp}_{SO(n)}: \mathbb{R}^{\frac{n(n-1)}{2}} \mapsto SO(n), \quad \text{Exp}_{SO(n)}(\boldsymbol{\delta}) = \exp\left([\boldsymbol{\delta}]\right) \\
            &\text{Log}_{SO(n)}:  SO(n) \mapsto \mathbb{R}^{\frac{n(n-1)}{2}} , \quad \text{Log}_{SO(n)}(\mathbf{x}) =  \log\left(\mathbf{x}\right)^\vee
        \end{align}
        \setlength{\abovedisplayskip}{0.15cm} 
        \setlength{\belowdisplayskip}{0.15cm}
    \end{subequations}
    for $n = 2$,
    \begin{eqnarray}
        \text{Exp}\left(\delta\right) = \begin{bmatrix}
            \cos \delta & -\sin \delta \\ \sin \delta & \cos \delta
        \end{bmatrix} , \ \text{Log}(\mathbf{x}) = \text{atan2}(\mathbf{x}_{\text{21}}, \mathbf{x}_{\text{11}})
    \end{eqnarray}
    for $n = 3$,
    \begin{subequations}
        \small
        \begin{align}
            \text{Exp}\left(\boldsymbol{\delta}\right) &= \mathbf{I} + \frac{\lfloor\boldsymbol{\delta}\rfloor}{\Vert \boldsymbol{\delta} \Vert} \sin\Vert \boldsymbol{\delta} \Vert + \frac{\lfloor\boldsymbol{\delta}\rfloor^2}{\Vert \boldsymbol{\delta} \Vert}\left(1 - \cos\Vert \boldsymbol{\delta} \Vert\right) 
            \label{e:exp_SO3}\\
            \text{Log}\left(\mathbf{x}\right)^\vee &= \frac{\alpha \left(\mathbf{x} - \mathbf{x}^T\right)}{2\sin\alpha}, \quad \alpha = \cos^{-1}\left(\frac{\text{tr}(\mathbf{x}) - 1}{2}\right)
        \end{align}
        \setlength{\abovedisplayskip}{0.15cm} 
        \setlength{\belowdisplayskip}{0.15cm}
    \end{subequations}
    where the operations $\lfloor \cdot \rfloor$ takes a vector to form a skew-symmetric matrix, and $(\cdot)^\vee$ is the inverse map.
    
    \subsubsection{\textbf{\textit{Example 3: Two-Sphere $\mathbb{S}^2_r$}}}~
    
    A vector in $\mathbb{R}^3$ with fixed magnitude, e.g., pendulum movement in robotic systems, is on the manifold of two-sphere with dimension of $2$: $\mathbb{S}^2_r \triangleq \{\mathbf{x}\in\mathbb{R}^3\ | \ \Vert\mathbf{x}\Vert = r, r > 0\}$. Given two vectors $\mathbf{x}, \mathbf{y}$ on $\mathbb{S}^2_r$, the perturbation between them can be described as a rotation that steers $\mathbf x$ to $\mathbf y$ along the shortest path (i.e., {\it geodesic}). The shortest rotation is always on the tangent plane of $\mathbf x$, hence has two degree of freedom. Let $\mathbf{b}_1, \mathbf{b}_2 \in \mathbb{R}^3$ be two orthonormal basis vectors of the tangent plane of $\mathbf x$ and $\mathbf{B}(\mathbf x) = [\mathbf{b}_1 \  \mathbf{b}_2]$, the $\boxminus$ is then \cite{he2021embedding}:
    \begin{eqnarray}
        \mathbf{y} \boxminus \mathbf{x} = \mathbf{B}(\mathbf{x})^T \left(\theta \frac{\lfloor\mathbf{x}\rfloor\mathbf{y}}{\Vert \lfloor\mathbf{x}\rfloor\mathbf{y} \Vert} \right), \ \theta = \text{atan2}(\Vert \lfloor\mathbf{x}\rfloor\mathbf{y} \Vert, \mathbf{x}^T\mathbf{y})
    \end{eqnarray}
    
    There is no unique way to determine the basis vector $\mathbf{B}(\mathbf x)$, one simple method adopted by \cite{he2021embedding} is to align one of the natural coordinate basis $\mathbf{e}_1, \mathbf{e}_2, \mathbf{e}_3$ with $\mathbf{x}$, and the rest two basis would be $\mathbf{b}_1$ and $\mathbf{b}_2$. 

    With the axiom $\mathbf{x} \boxplus (\mathbf{y} \boxminus \mathbf{x}) = \mathbf{y}$ and the fact $\mathbf{B}^T(\mathbf{x})\mathbf{B}(\mathbf{x}) = \mathbf{I}_{2\times 2}$, we have:
    \begin{eqnarray}
        \mathbf{x} \boxplus \boldsymbol{\delta} = \text{Exp}\left(\mathbf{B}(\mathbf{x}) \boldsymbol{\delta}\right)\cdot \mathbf{x}, \quad \boldsymbol{\delta} \in \mathbb{R}^2
    \end{eqnarray}
    
    
    Practical systems evolving on $\mathbb{S}^2$ are usually driven by exogenous rotations, which do not necessarily lie in the tangent plane defining $\boxplus$. Hence, the $\oplus$ should be defined separately, as below:
    
    \begin{equation}
        \small
        \mathbf{x} \oplus \boldsymbol{\delta}^e = \text{Exp}(\boldsymbol{\delta}^e) \cdot \mathbf{x} , \quad  \boldsymbol{\delta}^e \in \mathbb{R}^3
        \setlength{\abovedisplayskip}{0.15cm} 
        \setlength{\belowdisplayskip}{0.15cm}
    \end{equation} 
    
    \subsubsection{\textbf{\textit{Example 4: 2D Surface $\mathcal{S}$}}} ~
    
    \begin{figure}[!t]
        \centering
        \includegraphics[width=0.5\linewidth]{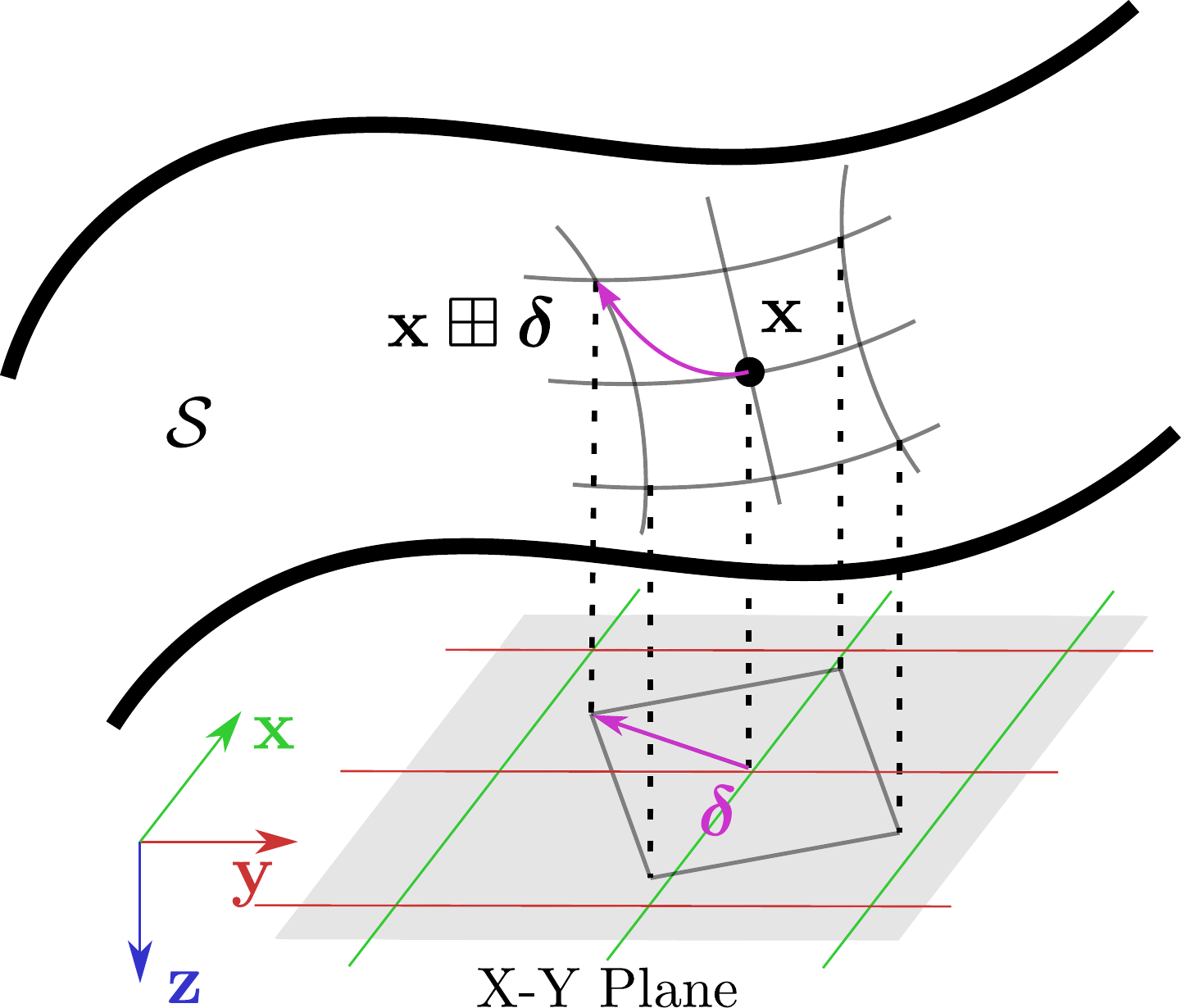} 
        \caption{Illustration of the $\boxplus$ operations on 2D surface $S$ with a perturbation in the homoemorphic space, the $X$-$Y$ plane.}
        \label{f:surface}
        \setlength{\abovecaptionskip}{-0.0cm} 
        \setlength{\belowcaptionskip}{-0.2cm}
        \vspace{-0.3cm}
    \end{figure}
    
    Consider a ground vehicle moving on ground surface, which is smooth but not always flat. Such a ground surface can be generally modeled as a smooth surface manifold embedded in $\mathbb{R}^3$: $\mathcal{S} \triangleq \{ \mathbf x \in \mathbb{R}^3 | z = F(x, y)\}$, where $z = F(x,y)$ is the height parameterized by horizontal coordinates $(x, y)$ \cite{zhang2021pose}. Unlike previous examples where the perturbations of $\boxplus$ are represented in the tangent plane, for the 2D surface $\mathcal{S}$, the perturbation $\boldsymbol{\delta} \in \mathbb{R}^2$ can be directly represented in the $X$-$Y$ plane of the space that embeds $\mathcal{S}$ (see Fig. \ref{f:surface}), leading to
    
    \begin{equation}
        \small
            \mathbf{x} \boxplus \boldsymbol{\delta} = \begin{bmatrix} \mathbf{E}_{12}^T \mathbf{x} + \boldsymbol{\delta} \\ F\left(\mathbf{E}_{12}^T \mathbf{x} + \boldsymbol{\delta}\right)
            \end{bmatrix} , \quad \mathbf{y} \boxminus \mathbf{x} = \mathbf{E}_{12}^T \left(\mathbf{y} - \mathbf{x}\right)
        \setlength{\abovedisplayskip}{0.15cm} 
        \setlength{\belowdisplayskip}{0.15cm}
    \end{equation}
    where $\mathbf{E}_{12} = [\mathbf{e}_1 \  \mathbf{e}_2] \in \mathbb{R}^{3 \times 2}$.
    
    
    Since the actuator of a real ground vehicle (e.g., steering, acceleration) usually can only drive the robot along the ground surface, the exogenous perturbation lies one the tangent plane of the robot's current position. However, such exogenous perturbation in the tangent plane can always be projected to the $x-y$ plane and hence makes use of $\boxplus$ to describe the effect, meaning $\oplus \simeq \boxplus$. We will show how to achieve such projection in the example system in Section. \ref{sec:groundvehicle}.

    \subsubsection{\textbf{\textit{Example 5: Compound Manifolds}}} ~
    
    The cartesian product \cite{hertzberg2013integrating} of two (or several) manifolds $\mathcal{M}_1, \mathcal{M}_2$ is considered as a manifold $\mathcal{M} = \mathcal{M}_1 \times \mathcal{M}_2$, and satisfies 
    \begin{equation}
        \label{e:compound}
        \begin{pmatrix}
            \mathbf{x}_1 \\ \mathbf{x}_2
        \end{pmatrix} \boxplus \begin{bmatrix}
            \boldsymbol{\delta}_1 \\ \boldsymbol{\delta}_2
        \end{bmatrix} = \begin{pmatrix}
            \mathbf{x}_1 \boxplus \boldsymbol{\delta}_1 \\ \mathbf{x}_2 \boxplus \boldsymbol{\delta}_2
        \end{pmatrix},\ \begin{pmatrix}
        \mathbf{x}_1 \\ \mathbf{x}_2
        \end{pmatrix} \oplus \begin{bmatrix}
        \boldsymbol{\delta}^e_1 \\ \boldsymbol{\delta}^e_2
        \end{bmatrix} = \begin{pmatrix}
        \mathbf{x}_1 \oplus \boldsymbol{\delta}^e_1 \\ \mathbf{x}_2 \oplus \boldsymbol{\delta}^e_2
        \end{pmatrix}
        \setlength{\abovedisplayskip}{0.15cm} 
        \setlength{\belowdisplayskip}{0.15cm}
    \end{equation}
    for $\mathbf{x}_1 \in \mathcal{M}_1$, $\mathbf{x}_2 \in \mathcal{M}_2$ , and $\boldsymbol{\delta}_1 \in \mathbb{R}^{n_1}$, $\boldsymbol{\delta}_2 \in \mathbb{R}^{n_2}$, $\boldsymbol{\delta}^e_1 \in \mathbb{R}^{l_1}$, $\boldsymbol{\delta}^e_2 \in \mathbb{R}^{l_2}$.
    

    \section{Symbolic MPC Formulation on Manifolds}
    \label{sec:canonical_representation}
    
    In this section, we propose a formal symbolic MPC formulation for trajectory tracking with a canonical representation of on-manifold systems.
    
    
   
    
    \subsection{Canonical Representation of On-Manifold Systems }
    Consider a robotic system operating on a compound manifold $\mathcal{M} = \mathcal{M}_1 \times \cdots \times \mathcal{M}_i$ with dimension of $n$, assume the exogenous perturbation caused by control input is constant for one sampling period $\Delta t$ (i.e., zero-order holder discretization), we can cast it into the following canonical form:
    \begin{eqnarray}
        \mathbf{x}_{k+1} = \mathbf{x}_k \oplus \Delta t \mathbf{f}(\mathbf{x}_k, \mathbf{u}_k)
        \label{e:original_sys}
        \setlength{\abovedisplayskip}{0.15cm} 
        \setlength{\belowdisplayskip}{0.15cm} 
    \end{eqnarray}
    where $\mathbf{x} \in \mathcal{M}$ is the on-manifold system state, $\mathbf{u} \in \mathbb{R}^m$ is the control input in the actuator space, and $\mathbf{f}(\mathbf{x}_k, \mathbf{u}_k) \in \mathbb{R}^n$ is the system-specific representation of the perturbation referred to Section. \ref{sec:operations}. For example, for attitude kinematics $\dot{\mathbf R} = \mathbf R \lfloor \boldsymbol{\omega} \rfloor$, we can discretize it into $\mathbf R_{k+1} = \mathbf R_k {\rm Exp}(\boldsymbol{\omega}_k \Delta t) = \mathbf R_k \oplus_{SO(3)} (\boldsymbol{\omega}_k \Delta t)$ assuming that the angular velocity $\boldsymbol{\omega}_k$ is constant over $\Delta t$. Similarly, for a 3D pendulum with position $\mathbf q \in \mathbb{S}^2$ and angular velocity $\boldsymbol{\omega}$, it satisfies $\mathbf q = \boldsymbol{\omega} \times \mathbf q$ \cite{wu2015variation}, which can be discretized as $\mathbf q_{k+1} = {\rm Exp}(\boldsymbol{\omega}_k \Delta t) \mathbf q_k = \mathbf q_{k} \oplus_{\mathbb{S}^2} (\boldsymbol{\omega}_k \Delta t)$. We refer readers to \cite{he2021embedding} for more examples of how to represent a system into the canonical form in  (\ref{e:original_sys}). 
    
    \subsection{On-Manifold MPC Formulation for Trajectory Tracking}
    
    Based on the canonical system representation in (\ref{e:original_sys}) and further make use of the $\boxminus$ to minimally represent the deviation between the actual state and trajectory, we can formally formulate the MPC problem for systems evolving on manifold, as follows:
    
    \begin{equation}
        \small
        \begin{aligned}
            &\displaystyle\min_{\mathbf{u}_0, \cdots, \mathbf{u}_{N-1}} \sum_{k=0}^{N-1} \left(\Vert \mathbf{x}_k \boxminus \mathbf{x}_k^d\Vert^2_{\mathbf{Q}_k} + \Vert \mathbf{u}_k \boxminus \mathbf{u}_k^d\Vert^2_{\mathbf{R}_k}\right) + \Vert \mathbf{x}_N \boxminus \mathbf{x}_N^d\Vert^2_{\mathbf{P}_N} \\
            &\  \mathrm{s.t.} \quad \mathbf{x}_{k+1} = \mathbf{x}_k \oplus \Delta t \mathbf{f}(\mathbf{x}_k, \mathbf{u}_k), \quad \mathbf x_0 = \mathbf x_{\text{init}} \\
            &\  \qquad \ 
            \mathbf{u}_{k} \in \mathbb{U} , \quad k = 0, \cdots , N-1
        \end{aligned}  
        \label{e:canonical_mpc}
        \setlength{\abovedisplayskip}{0.15cm} 
        \setlength{\belowdisplayskip}{0.15cm} 
    \end{equation}
    where the superscript $(\cdot)^d$ denotes the reference trajectory satisfying $\mathbf{x}_{k+1}^d = \mathbf{x}_k^d \oplus \Delta t \mathbf{f}(\mathbf{x}_k^d, \mathbf{u}_k^d)$ and $\mathbf{u}_k^d \in \mathbb{U}$. $\mathbf{Q}_k>0$ and $\mathbf{R}_k>0$ are respectively the stage state and input penalty matrices, while $\mathbf{P}_N>0$ is the penalty of the terminal state. $\boxminus$ and $\oplus$ operations are defined on the manifold $\mathcal{M}$ where the system state lies on. $\mathbb{U}$ is the boundary of inputs:
    \begin{equation}
        \small
        \mathbb{U}= \{\mathbf{u}\in \mathbb{R}^{m} | \mathbf{u}_{\min} \leq \mathbf{u} \leq \mathbf{u}_{\max}\}
        \setlength{\abovedisplayskip}{0.15cm} 
        \setlength{\belowdisplayskip}{0.15cm}
    \end{equation}

    \section{On-Manifold MPC Solution Based on the Error System}
    \label{sec:mpc}
    
\begin{figure}[!t]
        \centering
        \includegraphics[width=0.8\linewidth]{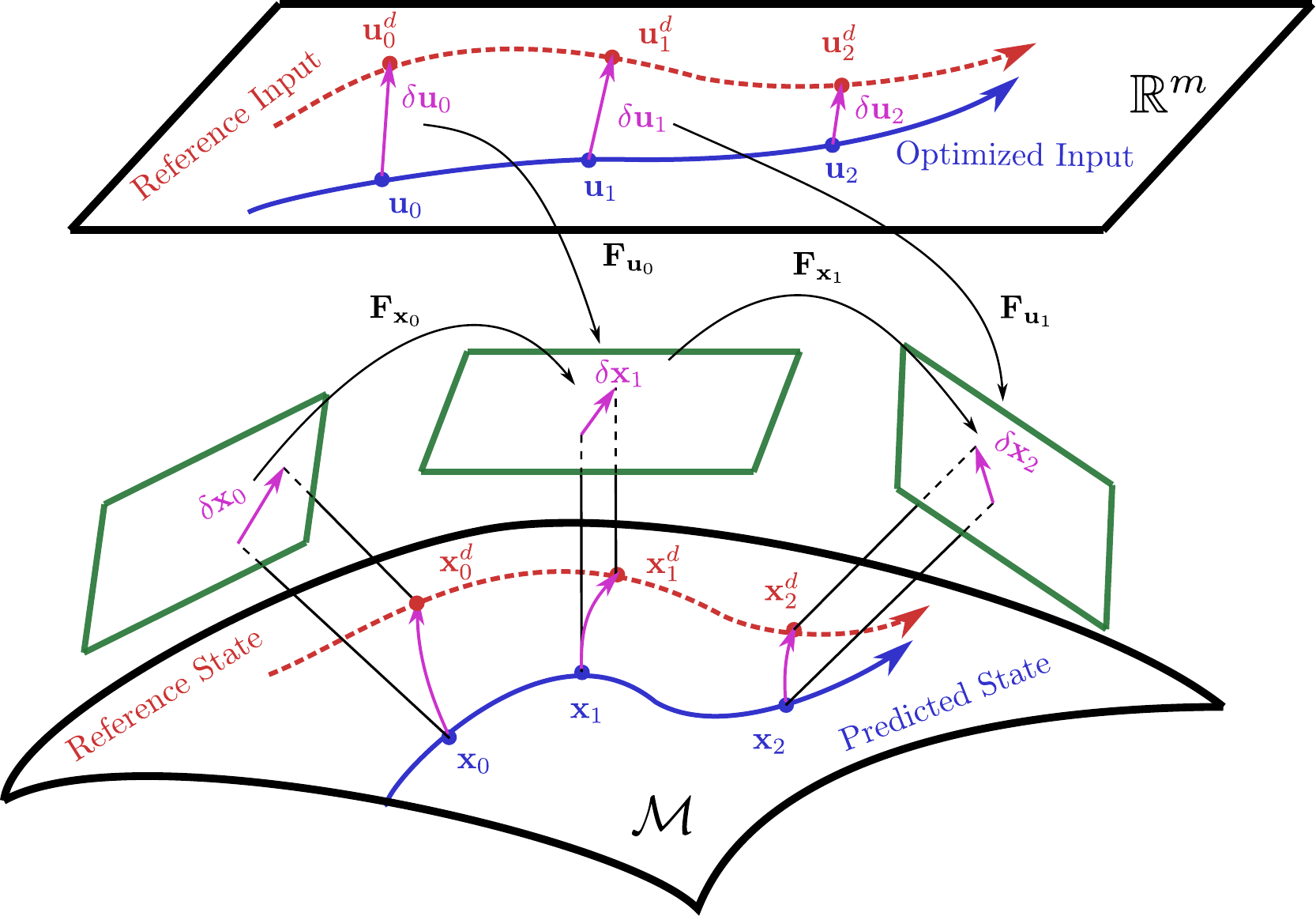} 
        \caption{Interpretation of the linearized error system}
        \label{f:mpc}
        \setlength{\abovecaptionskip}{-0.1cm} 
        \setlength{\belowcaptionskip}{-0.2cm}
        \vspace{-0.3cm}
    \end{figure}

    A standard MPC in Euclidean space for trajectory tracking works by solving a constraint optimization problem that minimizes both the state and input error while satisfies the system model and the input boundary. However, for the on-manifold MPC formulation in (\ref{e:canonical_mpc}), the manifold constraints that $\mathbf x_k \in \mathcal{M}$ must always be satisfied during the optimization. In this paper, we cope with the manifold constraints by considering an error system around the reference trajectory. On account of the introduced error-state system representation, the on-manifold state and input error between the reference and actual trajectory, are presented in Euclidean space that can be directly used in the MPC formulation (see Fig. \ref{f:mpc}). 
    
    \subsection{The Linearized Error System}
    
    The original on-manifold system dynamics (\ref{e:original_sys}) is nonlinear and with state representation by manifolds. Motivated by applying advanced linear control techniques (e.g. LQR and MPC) on such systems, the error-state system is derived from the original system (\ref{e:original_sys}). The error-state system is parameterized based on the perturbation operation of generic manifolds in Section \ref{sec:operations}, as follows:
    \begin{eqnarray}
        \delta \mathbf{x}_k &=& \mathbf{x}_k \boxminus \mathbf{x}_k^d  \in \mathbb{R}^n
        \label{e:error_state}\\
        \delta \mathbf{u}_k &=& \mathbf{u}_k \boxminus \mathbf{u}_k^d \in \mathbb{R}^m
        \label{e:error_input}
    \end{eqnarray}
    
    Substituting (\ref{e:error_state}) and (\ref{e:error_input}) into (\ref{e:original_sys}) leads to
    \begin{eqnarray}
        \delta \mathbf{x}_{k+1} &=& \left( \mathbf{x}_k \oplus \Delta t \mathbf{f}(\mathbf{x}_k, \mathbf{u}_k) \right) \boxminus \left( \mathbf{x}_k^d \oplus \Delta t \mathbf{f}(\mathbf{x}_k^d, \mathbf{u}_k^d) \right) \nonumber \\
        &=& \left( \left(\mathbf{x}_k^d \boxplus \delta\mathbf{x}_k\right) \oplus \Delta t \mathbf{f}(\mathbf{x}_k^d \boxplus \delta\mathbf{x}_k, \mathbf{u}_k^d \boxplus \delta\mathbf{u}_k)\right)  \nonumber \\
        &\ & \quad \boxminus \left( \mathbf{x}_k^d \oplus \Delta t \mathbf{f}(\mathbf{x}_k^d, \mathbf{u}_k^d) \right)
        \label{e:original_error_system}
    \end{eqnarray}
    
    Although the error system in (\ref{e:original_error_system}) is represented by the minimal parametrization $\delta \mathbf{x}_k$ and $\delta \mathbf{u}_k$, it suffers from no singularity because the tracking controller usually keeps the state error small. The minimally parameterized error system is equivalent to the original on-manifold system, but is now a usual nonlinear system operating in Euclidean space $\mathbb{R}^n$. Along with the objective function in (\ref{e:canonical_mpc}) which are also in terms of $\delta \mathbf{x}_k$ and $\delta \mathbf{u}_k$, they form a standard MPC problem in Euclidean space that can be solved by common MPC optimization techniques. Specifically, in this paper, we assume a small tracking error and linearize the error system and objective function around zero (i.e., $\delta \mathbf x_k \approx \mathbf 0$). This leads to a QP problem that can be solved efficiently. 
    
    Performing Taylor series expansion to (\ref{e:original_error_system}) and keeping terms up to the first order lead to the linearization below:
    \begin{eqnarray}
        \delta \mathbf{x}_{k+1} \approx \mathbf{F}_{\mathbf{x}_k} \delta\mathbf{x}_k + \mathbf{F}_{\mathbf{u}_k} \delta\mathbf{u}_k 
        \label{e:errot_state_system}
    \end{eqnarray}
    where $\mathbf{F}_{\mathbf{x}_k}$ is the partial differentiation with respect to (w.r.t.) $\delta \mathbf{x}_k$ (evaluated at zero) and can be computed by the chain rule as below:
    \begin{equation}
        \small
        \begin{aligned}
            \mathbf{F}_{\mathbf{x}_k} &= \frac{\partial \left( \left( (\mathbf{x}_k^d \boxplus \delta\mathbf{x}_k ) \oplus \Delta t \mathbf{f}(\mathbf{x}_k^d , \mathbf{u}_k^d )\right) \boxminus \left( \mathbf{x}_k^d \oplus \Delta t \mathbf{f}(\mathbf{x}_k^d, \mathbf{u}_k^d) \right) \right)}{\partial \delta\mathbf{x}_k} \\
            & + \frac{\partial \left( \left( \mathbf{x}_k^d  \oplus \Delta t \mathbf{f}(\mathbf{x}_k^d\boxplus \delta\mathbf{x}_k , \mathbf{u}_k^d )\right) \boxminus \left( \mathbf{x}_k^d \oplus \Delta t \mathbf{f}(\mathbf{x}_k^d, \mathbf{u}_k^d) \right) \right)}{\partial \Delta t \mathbf{f}(\mathbf{x}_k^d\boxplus \delta\mathbf{x}_k , \mathbf{u}_k^d )}\\
            & \quad \cdot \Delta t \frac{\partial \mathbf{f}(\mathbf{x}_k^d\boxplus \delta\mathbf{x}_k , \mathbf{u}_k^d )}{\partial \delta\mathbf{x}_k} \bigg|_{\delta\mathbf{x}_k = 0}\\
            &= \mathbf{G}_{\mathbf{x}_k} + \Delta t \mathbf{G}_{\mathbf{f}_k} \cdot \frac{\partial \mathbf{f}(\mathbf{x}_k^d\boxplus \delta\mathbf{x}_k , \mathbf{u}_k^d )}{\partial \delta\mathbf{x}_k} \bigg|_{\delta\mathbf{x}_k = 0}
        \end{aligned}  
        \label{e:F_x}
        \setlength{\abovedisplayskip}{0.15cm} 
        \setlength{\belowdisplayskip}{0.15cm}
    \end{equation}
    and $\mathbf{F}_{\mathbf{u}_k}$ is the partial differentiation w.r.t. $\delta \mathbf{u}_k$ (evaluated at zero)
    \begin{equation}
        \small
        \begin{aligned}
            \mathbf{F}_{\mathbf{u}_k} &= \frac{\partial \left( \left( \mathbf{x}_k^d \oplus \Delta t \mathbf{f}(\mathbf{x}_k^d , \mathbf{u}_k^d \boxplus \delta\mathbf{u}_k)\right) \boxminus \left( \mathbf{x}_k^d \oplus \Delta t \mathbf{f}(\mathbf{x}_k^d, \mathbf{u}_k^d) \right) \right)}{\partial \Delta t\mathbf{f}(\mathbf{x}_k^d , \mathbf{u}_k^d \boxplus \delta\mathbf{u}_k)} \\
            & \quad \cdot \Delta t \frac{\partial \mathbf{f}(\mathbf{x}_k^d , \mathbf{u}_k^d \boxplus \delta\mathbf{u}_k)}{\partial \delta\mathbf{u}_k} \bigg|_{\delta\mathbf{u}_k = 0}\\
            &= \Delta t \mathbf{G}_{\mathbf{f}_k} \frac{\partial \mathbf{f}(\mathbf{x}_k^d , \mathbf{u}_k^d \boxplus \delta\mathbf{u}_k)}{\partial \delta\mathbf{u}_k} \bigg|_{\delta\mathbf{u}_k = 0}
        \end{aligned}    
        \label{e:F_u}
        \setlength{\abovedisplayskip}{0.15cm} 
        \setlength{\belowdisplayskip}{0.15cm}
    \end{equation}
    where 
    \begin{equation}
    \small \label{Gx_Gf}
    \begin{aligned}
    \mathbf{G}_{{\mathbf{x}}_{k} } &\!= \!\left. \frac{\partial \!\left( \!\left(\! (\mathbf{x}_k^d \!\boxplus \!\boldsymbol{\delta} ) \!\oplus \!\Delta t \mathbf{f}(\mathbf{x}_k^d , \mathbf{u}_k^d ) \! \right) \!\boxminus \!\left( \mathbf{x}_k^d \!\oplus \!\Delta t \mathbf{f}(\mathbf{x}_k^d, \mathbf{u}_k^d) \right)\! \right)}{\partial \boldsymbol{\delta}} \right|_{\substack{\boldsymbol{\delta} = \mathbf 0}} \\
    \mathbf{G}_{{\mathbf{f}}_{\tau} } &\!=\!\left. \frac{\partial \!\left( \!\left( \mathbf{x}_k^d  \!\oplus \!\left( \Delta t \mathbf{f}(\mathbf{x}_k^d, \mathbf{u}_k^d ) \!+ \!\boldsymbol{\delta} \right) \! \right) \!\boxminus\! \left( \mathbf{x}_k^d \!\oplus \!\Delta t \mathbf{f}(\mathbf{x}_k^d, \mathbf{u}_k^d) \right)\! \right)}{\partial \boldsymbol{\delta} } \right|_{\substack{\boldsymbol{\delta} = \mathbf 0
    }}
    \end{aligned}
    \setlength{\abovedisplayskip}{0.15cm} 
    \setlength{\belowdisplayskip}{0.15cm}
\end{equation}
    
    
    \subsection{Error-State MPC Formulation}
    
    Based on the linearized system (\ref{e:errot_state_system}), the original MPC problem in (\ref{e:canonical_mpc}) can be transformed into
    
    \begin{equation}
        \small
        \begin{aligned}
            &\min_{\delta\mathbf{u}_0, \cdots, \delta\mathbf{u}_{N-1}} \sum_{k=0}^{N-1} \left(\Vert \delta\mathbf{x}_k\Vert^2_{\mathbf{Q}_k} + \Vert\delta\mathbf{u}_k)\Vert^2_{\mathbf{R}_k}\right) + \Vert \delta\mathbf{x}_N\Vert^2_{\mathbf{P}_N} \\
            &\  \mathrm{s.t.} \quad \delta \mathbf{x}_{k+1} = \mathbf{F}_{\mathbf{x}} \delta\mathbf{x}_k + \mathbf{F}_{\mathbf{u}} \delta\mathbf{u}_k, \quad \delta\mathbf{x}_{0} = \delta\mathbf{x}_{\text{init}}\\ 
            &\  \qquad \ 
            \delta \mathbf{u}_{k} \in \delta \mathbb{U}_k, \quad k = 0, \cdots , N-1 
        \end{aligned}  
        \label{e:error_state_mpc}
        \setlength{\abovedisplayskip}{0.15cm} 
        \setlength{\belowdisplayskip}{0.15cm}
    \end{equation}
     where $\delta\mathbf{x}_{\text{init}} = \mathbf x_{\text{init}} \boxminus \mathbf x_0^d$ and $\delta \mathbb{U}_k = \{\delta \mathbf{u}\in \mathbb{R}^{m} | \mathbf{u}_{\min} - \mathbf u_k^d \leq \delta \mathbf{u} \leq \mathbf{u}_{\max} - \mathbf u_k^d \}$.

    \subsection{Optimization Solution}
    To efficiently solve the optimization problem in (\ref{e:error_state_mpc}), we propose an approach that parameterizes the state sequence in terms of the input vector to achieve a compact objective function. Instead of dealing with it as an optimization constraint, the system function is implicit in the compact objective function.
    
    With error-state system (\ref{e:errot_state_system}), the sequence of error states can be given by
    \begin{eqnarray}
        \delta\mathbf{X} = \mathbf{M}\delta \mathbf{U} + \mathbf{H} \delta\mathbf{x}_0
        \label{e:compact_err_state_func}
    \end{eqnarray}
    where $\delta\mathbf{X} = [\delta\mathbf{x}_1^T \  \cdots \  \delta\mathbf{x}_N^T]^T$, $\delta\mathbf{U} = [
        \delta\mathbf{u}_0^T \  \cdots \  \delta\mathbf{u}_{N-1}^T
    ]^T$ and
    
    \begin{equation}
        \footnotesize
        \mathbf{M} = \begin{bmatrix}
            \mathbf{F}_{\mathbf{u}_0} & \mathbf{0} & \cdots & \mathbf{0} \\
            \mathbf{F}_{\mathbf{x}_1}\mathbf{F}_{\mathbf{u}_0} & \mathbf{F}_{\mathbf{u}_1} & \cdots & \mathbf{0} \\
            \vdots & \vdots & \ddots & \vdots \\
            \left(\prod\limits_{i=N-1}^{1} \mathbf{F}_{\mathbf{x}_i} \right)\mathbf{F}_{\mathbf{u}_0} & \left(\prod\limits_{i=N-2}^{1} \mathbf{F}_{\mathbf{x}_i}\right)\mathbf{F}_{\mathbf{u}_1} & \cdots & \mathbf{F}_{\mathbf{u}_{N-1}}
        \end{bmatrix}
        \label{e:compact_M}
        \setlength{\abovedisplayskip}{0.15cm} 
        \setlength{\belowdisplayskip}{0.15cm}
    \end{equation}
    
    \begin{equation}
        \small
        \mathbf{H} = \begin{bmatrix}
            \mathbf{F}_{\mathbf{x}_0}^T & \cdots & \left(\prod\limits_{i=N-1}^{0}\mathbf{F}_{\mathbf{x}_i}\right)^T
        \end{bmatrix}^T
        \label{e:compact_H}
        \setlength{\abovedisplayskip}{0.15cm} 
        \setlength{\belowdisplayskip}{0.15cm}
    \end{equation}

    Substituting (\ref{e:compact_err_state_func}) into the objective function of (\ref{e:error_state_mpc}), the optimization problem becomes:
    \begin{equation}
        \small
        \begin{aligned}
            &\min_{\delta\mathbf{U}} \Vert \delta\mathbf{U}
            \Vert^2_{\mathbf{M}^T\bar{\mathbf{Q}}\mathbf{M}+\bar{\mathbf{R}}} + 2\delta\mathbf{x}_0^T\mathbf{H}^T\bar{\mathbf{Q}}\mathbf{M} \delta\mathbf{U}\\
            &\  \mathrm{s.t.} \quad 
            \delta\mathbf{U} \in \delta\bar{\mathbb{U}} \\
            &\  \qquad \ 
            \delta\mathbf{x}_{0} = \delta\mathbf{x}_{\text{init}}
        \end{aligned}  
        \label{e:compact_mpc}
    \end{equation}
    where $\| \mathbf x \|_{\mathbf A} = \mathbf x^T \mathbf A \mathbf x$, and
    \begin{subequations}
        \small
        \begin{align}
        \bar{\mathbf{Q}}_k &= \text{diag}\begin{bmatrix}
            \mathbf{Q}_1 & \cdots & \mathbf{Q}_{N-1} & \mathbf{P}_N
        \end{bmatrix} \\
        \bar{\mathbf{R}}_k &= \text{diag}\begin{bmatrix}
            \mathbf{R}_0 & \cdots & \mathbf{R}_{N-2}  & \mathbf{R}_{N-1}
        \end{bmatrix} \\ 
        \delta\bar{\mathbb{U}}_{\min} &= \begin{bmatrix}
            \mathbf{u}_{\min} - \mathbf{u}_{0}^d & \cdots & \mathbf{u}_{\min} - \mathbf{u}_{N-1}^d 
        \end{bmatrix} \\
        \delta\bar{\mathbb{U}}_{\max} &= \begin{bmatrix}
            \mathbf{u}_{\max} - \mathbf{u}_{0}^d & \cdots & \mathbf{u}_{\max} - \mathbf{u}_{N-1}^d 
        \end{bmatrix}
        \end{align} 
        \setlength{\abovedisplayskip}{0.15cm} 
        \setlength{\belowdisplayskip}{0.15cm}
    \end{subequations}
    
    It is seen that the compact form of the optimization problem (\ref{e:compact_mpc}) is a standard QP problem which can be solved efficiently by the existing QP solvers, such as the OOQP \cite{gertz2003object}. Let $\delta\mathbf{U}^*$ be the solution of (\ref{e:compact_mpc}) and $\delta\mathbf{u}^*_0$ denote the element at the timestamp of $k = 0$, then the optimal control is:
    \begin{equation}
        \small
        \mathbf{u}^*_0 = \delta\mathbf{u}^*_0 + \mathbf{u}_{0}^d
        \setlength{\abovedisplayskip}{0.15cm} 
        \setlength{\belowdisplayskip}{0.15cm}
    \end{equation}
    
    If the constraint of input boundary is not considered, or it is a finite-horizon discrete LQR for trajectory tracking, there is a closed-form solution of the unconstrained QP problem as follows:
    \begin{equation}
        \small
        \delta\mathbf{U}^* = - \left(\mathbf{M}^T\bar{\mathbf{Q}}\mathbf{M} + \bar{\mathbf{R}}\right)^{-1}  \mathbf{M}^T\bar{\mathbf{Q}}\mathbf{H} \delta\mathbf{x}_0
        \setlength{\abovedisplayskip}{0.15cm} 
        \setlength{\belowdisplayskip}{0.15cm}
    \end{equation}

    \subsection{On-manifold MPC}
    The procedure of the proposed error-state MPC for trajectory tracking on manifolds is summarized in Algorithm \ref{a:mpc}. As can be seen, the MPC algorithm iterates only once around the reference trajectory. In case where the state trajectory is not available, we could adopt Algorithm \ref{a:mpc} to produce a feasible trajectory naturally on the manifold via an iterative LQR style \cite{abbeel2010autonomous}: starting from an initial feasible trajectory (e.g., generated by a random sequence of input), the error system around the trajectory could be obtained and linearized to solve for the optimal control via Algorithm \ref{a:mpc}. The optimal control is then rolled out to obtain an updated trajectory, which is then used for the next iteration until convergence. 
    
    \begin{algorithm}[t]
        \SetAlgoLined
        \textbf{Initialization: } $k = 0$ \\
        \textbf{Given: } $N, \mathbf{u}_{\min}, \mathbf{u}_{\max}, \bar{\mathbf{Q}}, \bar{\mathbf{R}}$ \\    
        \While{\rm ControllerIsRunning}{
            $\mathbf{x}_0 \leftarrow$  StateEstimation $\mathbf{x}_k$\\
            $\mathbf{X}_{d_{0\to N-1}} \leftarrow$  Reference.State $\mathbf{x}_k^d, \cdots, \mathbf{x}_{k+N-1}^d$ \\
            $\mathbf{U}_{d_{0\to N-1}} \leftarrow$  Reference.Input $\mathbf{u}_k^d, \cdots, \mathbf{u}_{k+N-1}^d$ \\
            $\delta \mathbf{x}_0 \leftarrow$ GetCurrentStateError: $\delta \mathbf{x}_k = \mathbf{x}_k \boxminus \mathbf{x}_k^d$ \\
            Compute $\delta\bar{\mathbb{U}}_{\min}, \delta\bar{\mathbb{U}}_{\max}$ \\
            Compute $\mathbf{M}, \mathbf{H}$ \\
            $\delta \mathbf{U}^* \leftarrow$ Solve QP (\ref{e:compact_mpc})  \\
            $\delta \mathbf{u}_0^* \leftarrow$ ExtractFirstInput($\delta \mathbf{U}^*$) \\
            $\mathbf{u}_0^* = \delta\mathbf{u}^*_0 + \mathbf{u}_{0}^d$ \\
            ApplyInput $\mathbf{u}_0^*$ \\
            $k = k +1$
        }
        \caption{MPC for On-manifold Trajectory Tracking}
        \label{a:mpc}
    \end{algorithm}

\section{Isolation of Manifold Constraints}

 \begin{table*}[t]
        \centering
        \begin{tabular}{|l l l l l l|}
            \hline
            $\mathcal{M}$ & $\mathbf{x} \boxplus \boldsymbol{\delta}$ & $\mathbf{y} \boxminus \mathbf{x}$ & $\mathbf{x} \oplus \boldsymbol{\delta}^e$ & $\mathbf{G}_{\mathbf{x}_k}$ & $\mathbf{G}_{\mathbf{f}_k}$ \\
            \hline
            & & & & & \\[-1.7em]
            \hline
            $\mathbb{R}^n$ & $\mathbf{x} + \boldsymbol{\delta}$ & $\mathbf{y} - \mathbf{x}$& $\mathbf{x} + \boldsymbol{\delta}^e$ & $\mathbf{I}_{n}$ & $\mathbf{I}_{n}$\\
            \hline          
            $SO(2)$ & $\mathbf{x} \cdot  \text{Exp}(\boldsymbol{\delta})$ & $\text{Log}\left(\mathbf{x}^{-1}\mathbf{y}\right)$ & $\mathbf{x} \cdot  \text{Exp}(\boldsymbol{\delta}^e)$ &  $1$ & $1$ \\
            \hline
            $SO(3)$ &$\mathbf{x} \cdot  \text{Exp}(\boldsymbol{\delta})$ & $\text{Log}\left(\mathbf{x}^{-1}\mathbf{y}\right)$ & $\mathbf{x} \cdot  \text{Exp}(\boldsymbol{\delta}^e)$ & $\text{Exp}\left(-\Delta t \mathbf{f}_k \right)$ &$\mathbf{A}\left(\Delta t \mathbf{f}_k\right)^T$ \\
            \hline
            \multirow{2}{*}{$\mathbb{S}^2_r$} & \multirow{2}{*}{$\text{Exp}\left(\mathbf{B}(\mathbf{x}) \boldsymbol{\delta}\right) \cdot \mathbf{x}$} & \multirow{2}{*}{$\mathbf{B}(\mathbf{x})^T \left(\theta \frac{\lfloor\mathbf{x}\rfloor\mathbf{y}}{\Vert \lfloor\mathbf{x}\rfloor\mathbf{y} \Vert} \right)$} & \multirow{2}{*}{$\text{Exp}(\boldsymbol{\delta}^e) \cdot \mathbf{x}$} & $-\frac{1}{r^2}\mathbf{B}\left(\mathbf{x}_{k}^d \oplus \Delta t \mathbf{f}_k\right)^T $ & $-\frac{1}{r^2}\mathbf{B}\left(\mathbf{x}_{k}^d \oplus \Delta t \mathbf{f}_k\right)^T $\\
            &  & & & $ \quad \cdot \text{Exp}(\Delta t \mathbf{f}_k) \lfloor \mathbf{x}_k^d \rfloor^2 \mathbf{B}(\mathbf{x}_k^d)$ & $\quad \cdot \text{Exp}(\Delta t \mathbf{f}_k) \lfloor \mathbf{x}_k^d\rfloor^2 \mathbf{A}(\Delta t \mathbf{f}_k)^T$\\
            \hline
            \multirow{2}{*}{$\mathcal{S}$} & \multirow{2}{*}{$\begin{bmatrix} \mathbf{E}_{12}^T \mathbf{x} + \boldsymbol{\delta} \\ F\left(\mathbf{E}_{12}^T \mathbf{x} + \boldsymbol{\delta}\right)
            \end{bmatrix}$} & \multirow{2}{*}{$\mathbf{E}_{12}^T(\mathbf{y}-\mathbf{x})$} & \multirow{2}{*}{$\begin{bmatrix} \mathbf{E}_{12}^T \mathbf{x} + \boldsymbol{\delta}^e \\ F\left(\mathbf{E}_{12}^T \mathbf{x} + \boldsymbol{\delta}^e \right)
            \end{bmatrix}$} & $ \mathbf{I}_2  $ & \multirow{2}{*}{$\mathbf{I}_2$} \\
            & & &  &  & \\
            \hline 
        \end{tabular}
        \caption{Operations and manifold-specific parts of different primitive manifolds. $\mathbf{f}_k$ denotes $\mathbf{f}(\mathbf{x}_k^d, \mathbf{u}_k^d)$ for simplicity. Detail derivations and matrix $\mathbf{A}(\cdot)$ are given in the Appendix.}
        \label{tab_important_manifolds}
    \end{table*}
    
    It can be seen from the previous sections that in the on-manifold MPC formulation (\ref{e:canonical_mpc}), the system modeling,  error-system  derivation,  linearization and  optimization has the  manifold  constraints  completely decoupled  from  the  system dynamics. More specifically, the objective equation in (\ref{e:canonical_mpc}) consists only the manifold-specific operation $\boxminus$, the state equation in (\ref{e:canonical_mpc}) breaks into the manifold-specific operation $\oplus$ and system-specific part $\mathbf{f}\left(\mathbf{x}, \mathbf{u}\right)$, the two matrices $\mathbf{F}_{\mathbf{x}}$ in (\ref{e:F_x}) and $\mathbf{F}_{\mathbf{u}}$ in (\ref{e:F_u}) used in the error system linearization and optimization break into the manifold-specific parts $\mathbf{G}_{\mathbf{x}}$, $\mathbf{G}_{\mathbf{f}}$ and system-specific parts $ \frac{\partial\mathbf{f}\left(\mathbf{x} \boxplus \delta \mathbf{x},\mathbf{u}\right)}{\partial \delta \mathbf{x}}|_{\delta \mathbf x= \mathbf 0}, \frac{\partial\mathbf{f}\left(\mathbf{x},\mathbf{u} \boxplus \delta \mathbf u \right)}{\partial \delta \mathbf{u} }|_{ \delta \mathbf u= \mathbf 0}$. 
    
    The nice separation property between the manifold constraints and system dynamics allows the natural integration of manifold properties into the MPC framework, and only leaves system-specific parts (i.e., $\mathbf{f}\left(\mathbf{x}, \mathbf{u}\right)$, $ \frac{\partial\mathbf{f}\left(\mathbf{x} \boxplus \delta \mathbf{x},\mathbf{u}\right)}{\partial \delta \mathbf{x}}, \frac{\partial\mathbf{f}\left(\mathbf{x},\mathbf{u} \boxplus \delta \mathbf u \right)}{\partial \delta \mathbf{u} }$) to be filled for specific systems. Moreover, enabled by the manifold composition in (\ref{e:compound}), we only need to do so for simple {\it primitive manifolds} while those for larger {\it compound manifolds} can be automatically constructed by concatenating the {\it primitive manifolds}. The $\boxplus\backslash\boxminus$ and $\oplus$ operations for commonly used simple manifolds are summarized in Table. \ref{tab_important_manifolds} with detailed derivation shown in the Appendix. If a new primitive manifold is encountered, one could follow the procedures in Section. \ref{sec:operations} to define its $\boxplus\backslash\boxminus$ and $\oplus$ operations and the $\mathbf{G}_{\mathbf{x}}$ and $\mathbf{G}_{\mathbf{f}}$ defined in (\ref{Gx_Gf}). Note that these derivations are manifold-specific and are independent of system dynamics, hence need only be done once.

    \section{Applications and Experimental Results}
    \label{sec:experiment}  
    To verify the proposed MPC framework, we apply the canonical system representation and MPC implementation to two real-time robotic systems: a quadrotor UAV and an UGV moving on curved surface, and test the tracking performance.
    
    \begin{figure}[ht!]
    	\centering
    	\subfigure[] { \label{fig:quad_platform}     
			\includegraphics[width=0.45\columnwidth]{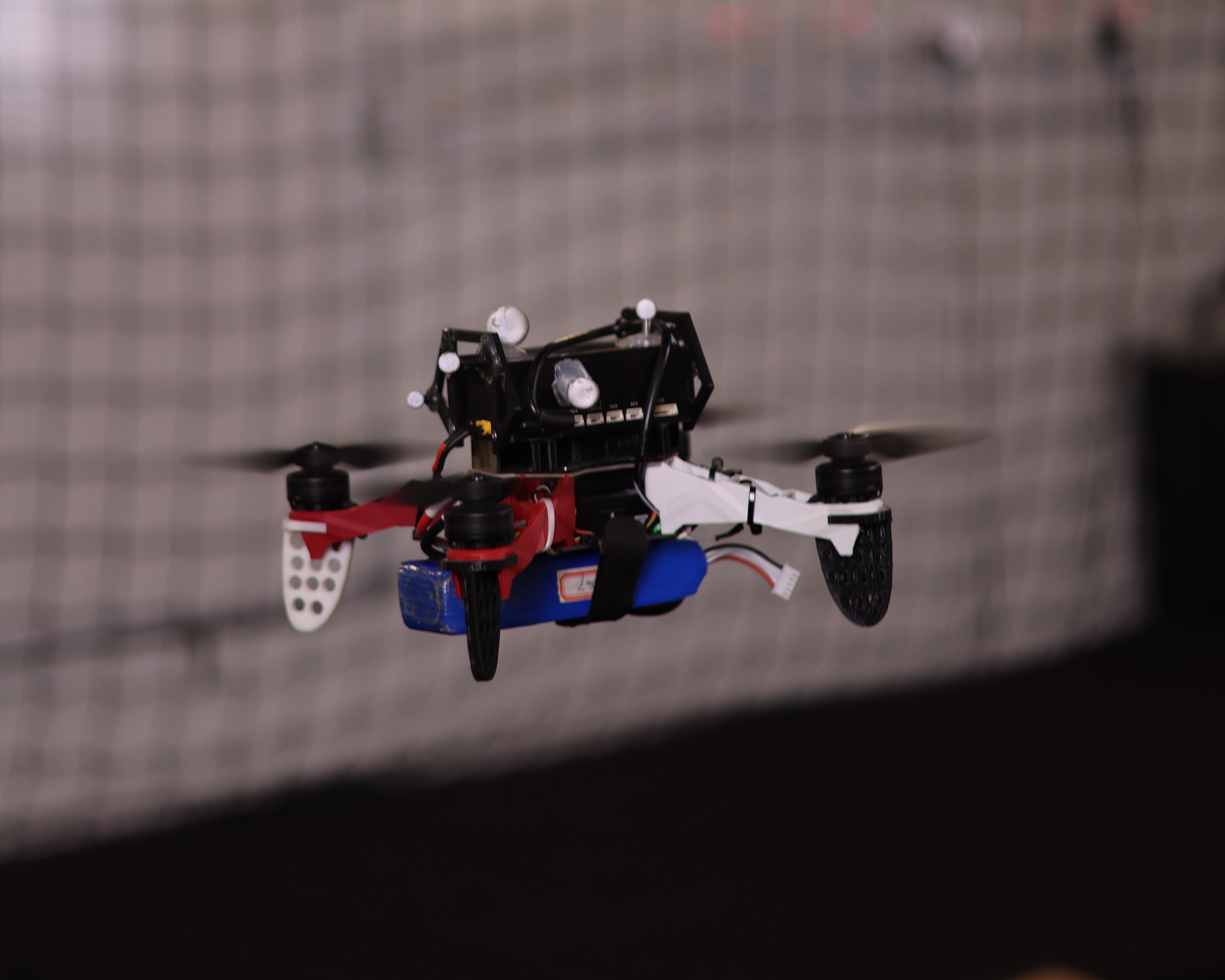}
		}
		\subfigure[] { \label{fig:car_platform}     
			\includegraphics[width=0.45\columnwidth]{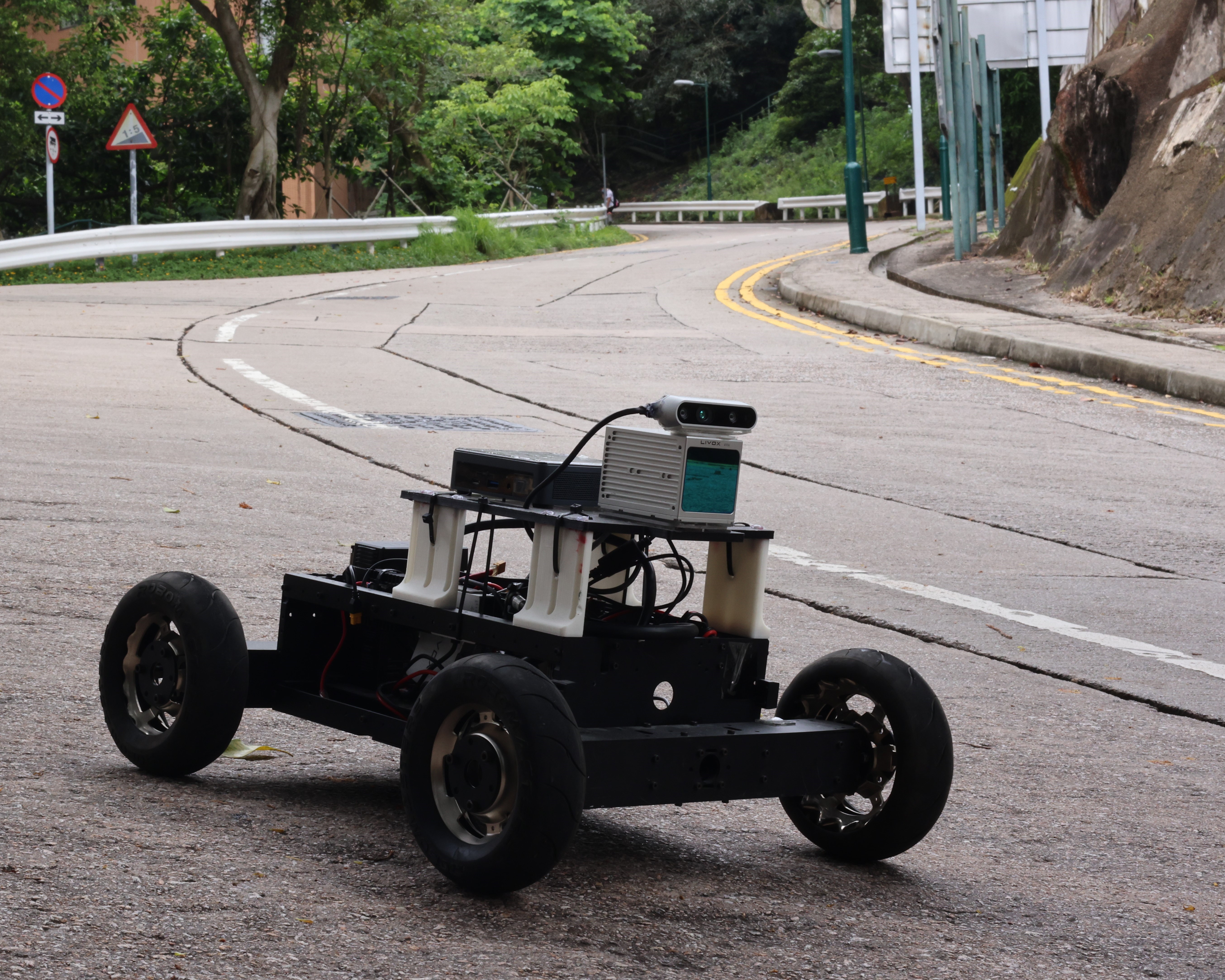}
		}
        \caption{Robotic platforms: quadrotor UAV and DJI RoboMaster UGV. }
        \setlength{\abovecaptionskip}{-0.1cm} 
        \setlength{\belowcaptionskip}{-0.2cm}
    	\label{fig:robotic_platform}
    \end{figure}
    
    \subsection{Application to Quadrotor UAVs}
    
    \begin{figure}[!ht]
        \centering
        \includegraphics[width=1\columnwidth]{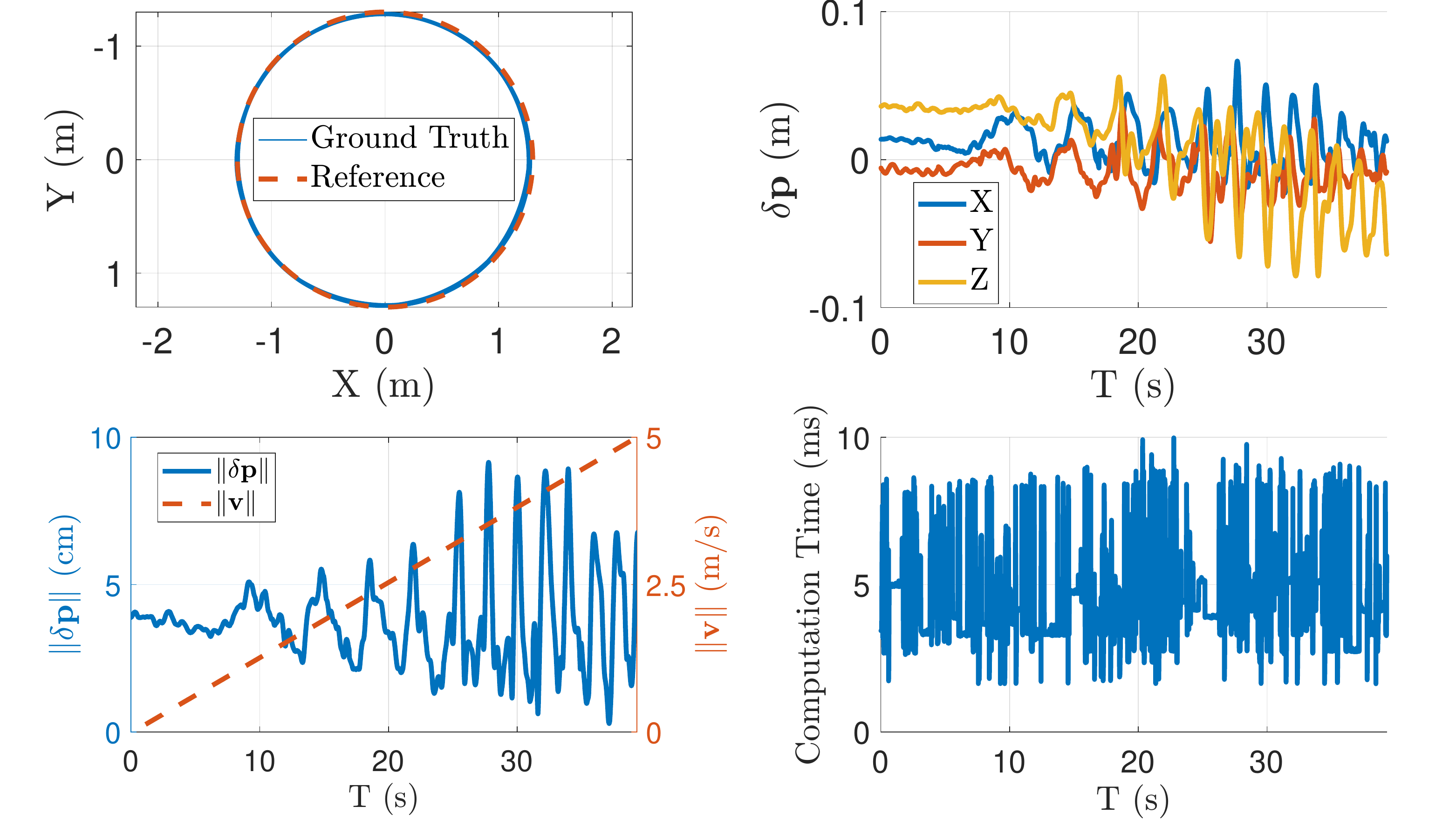}
        \caption{Control performance of the quadrotor tracks a horizontal circle trajectory with increasing speed  and maximal acceleration of $2g$}
        \label{fig:quad_horizontal_circle}
        \setlength{\abovecaptionskip}{-0.1cm} 
        \setlength{\belowcaptionskip}{-0.2cm}
        \vspace{-0.1in}
    \end{figure}
    
    \subsubsection{\textbf{\textit{Experiment Setup}}}
    The algorithm have been implemented and evaluated on the Q250 quadrotor frame (shown in Fig. \ref{fig:quad_platform}) equipped with PX4 mini, DJI Manifold-2C on-board computer with with a 1.8-4.0 GHz quad-core Intel i7-8550U CPU and 8GB RAM, DJI snail E100 electronic speed controller (ESC), motors and propellers. The vehicle state is estimated by Extented Kalman Filter (EKF) running on the autopilot with an external Optitrack motion capture system. Algorithms and communications are operated on Robot Operating System (ROS). 
    
    \subsubsection{\textbf{\textit{System Modeling}}}
    Let the subscript $\mathcal{I}$ and $\mathcal{B}$ respectively denote the world-fixed inertial frame and the body frame. The coordinate direction of $X$-$Y$-$Z$ is defined as forward-right-down. The vehicle motion is described by the position in the inertial frame $\mathbf{p}^{\mathcal{I}} \in \mathbb{R}^3$, the velocity in the inertial frame $\mathbf{v}^{\mathcal{I}} \in \mathbb{R}^3$, the orientation $\mathbf{R} \in SO(3)$ and the body angular rate $\boldsymbol{\omega}^{\mathcal{B}}$. 
    
    The formulation of the system model is written as:
    \begin{equation}
        \small
        \dot{\mathbf{p}}^{\mathcal{I}} = \mathbf{v}^{\mathcal{I}}, \quad \dot{\mathbf{v}}^{\mathcal{I}} = \mathbf{g}^{\mathcal{I}} - a_T\mathbf{R}\mathbf{e}_3, \quad \dot{\mathbf{R}} = \mathbf{R}\lfloor\boldsymbol{\omega}^{\mathcal{B}}\rfloor
        \label{e:quadrotor_model}
        \setlength{\abovedisplayskip}{0.15cm} 
        \setlength{\belowdisplayskip}{0.15cm}
    \end{equation}
    where $a_T$ denotes the scalar thrust acceleration and $\mathbf{g}^\mathcal{I}$ is the gravity vector with fixed length of $9.81m/s^2$.
    
    \subsubsection{\textbf{\textit{Canonical Representation}}}
    Based on the method of canonical system representation in Section \ref{sec:canonical_representation}, the quadrotor model in (\ref{e:quadrotor_model}) can be discretized and cast into the canonical form as follows:
    \begin{subequations}
        \small
        \begin{align}
            &\mathcal{M} = \mathbb{R}^3 \times \mathbb{R}^3 \times SO(3), \quad \dim(\mathcal{M}) = 9 \\
            &\mathbf{x} = \begin{pmatrix}
                \mathbf{p}^\mathcal{I} & \mathbf{v}^\mathcal{I} & \mathbf{R}
            \end{pmatrix} \in \mathcal{M},\quad  \mathbf{u} = \begin{bmatrix}
                a_T & \boldsymbol{\omega}^{\mathcal{B}}
            \end{bmatrix} \in \mathbb{R}^4 \\
            &\mathbf{f}(\mathbf{x}, \mathbf{u}) = \begin{bmatrix}
                \mathbf{v}^\mathcal{I} \\ \mathbf g^{\mathcal{I}} - a_{T}\mathbf{R}\mathbf{e}_3 \\ \boldsymbol{\omega}^{\mathcal{B}}
            \end{bmatrix} \in \mathbb{R}^6
        \end{align}
        \setlength{\abovedisplayskip}{0.15cm} 
        \setlength{\belowdisplayskip}{0.15cm}
    \end{subequations}
    and the system-specific parts:
    \begin{subequations}
        \small
        \begin{align}
            &\frac{\partial \mathbf{f}(\mathbf{x}\boxplus \delta\mathbf{x} , \mathbf{u} )}{\partial \delta\mathbf{x}} \bigg|_{\delta\mathbf{x} = \mathbf 0} =  \begin{bmatrix}
                \mathbf{0} & \mathbf{I}_3 & \mathbf{0} \\
                \mathbf{0} & \mathbf{0} & a_{T}\mathbf{R}\lfloor\mathbf{e}_3\rfloor \\
                \mathbf{0} & \mathbf{0} & \mathbf{0} 
            \end{bmatrix} \\
            &\frac{\partial \mathbf{f}(\mathbf{x} , \mathbf{u} \boxplus \delta\mathbf{u})}{\partial \delta\mathbf{u}} \bigg|_{\delta\mathbf{u} = \mathbf 0}  =  \begin{bmatrix}
                \mathbf{0} & \mathbf{0} \\
                -\mathbf{R}\mathbf{e}_3 & \mathbf{0} \\
                \mathbf{0} & \mathbf{I}_3
            \end{bmatrix}
        \end{align}
        \setlength{\abovedisplayskip}{0.15cm} 
        \setlength{\belowdisplayskip}{0.15cm}
    \end{subequations}

    \subsubsection{\textbf{\textit{Experimental Results}}}
    
     The proposed MPC runs on the on-board computer to calculate the target thrust and body angular rate at $100Hz$ with the predictive horizon of $8$,  while the body rate controller runs on PX4 at $250Hz$. The thrust command is computed by $T = k_T a_T$, where the thrust coefficient $k_T$ is assumed a positive constant such that $a_T = 9.8 m/s^2$ when the quadrotor is hovering.
     
     To demonstrate the tracking performance, we test the proposed MPC scheme on the quadrotor by flying along two aggressive maneuvers: a horizontal circle trajectory with linearly increasing speed up to $5 m/s$, and an acrobatic triple vertical flipping. Trajectories are designed offline and the quadrotor reference state and input along the trajectory are generated according to the quadtotor differential flatness \cite{mellinger2011minimum}.
     
\begin{figure}[!t] 
		\centering 
		\includegraphics[width=1\columnwidth]{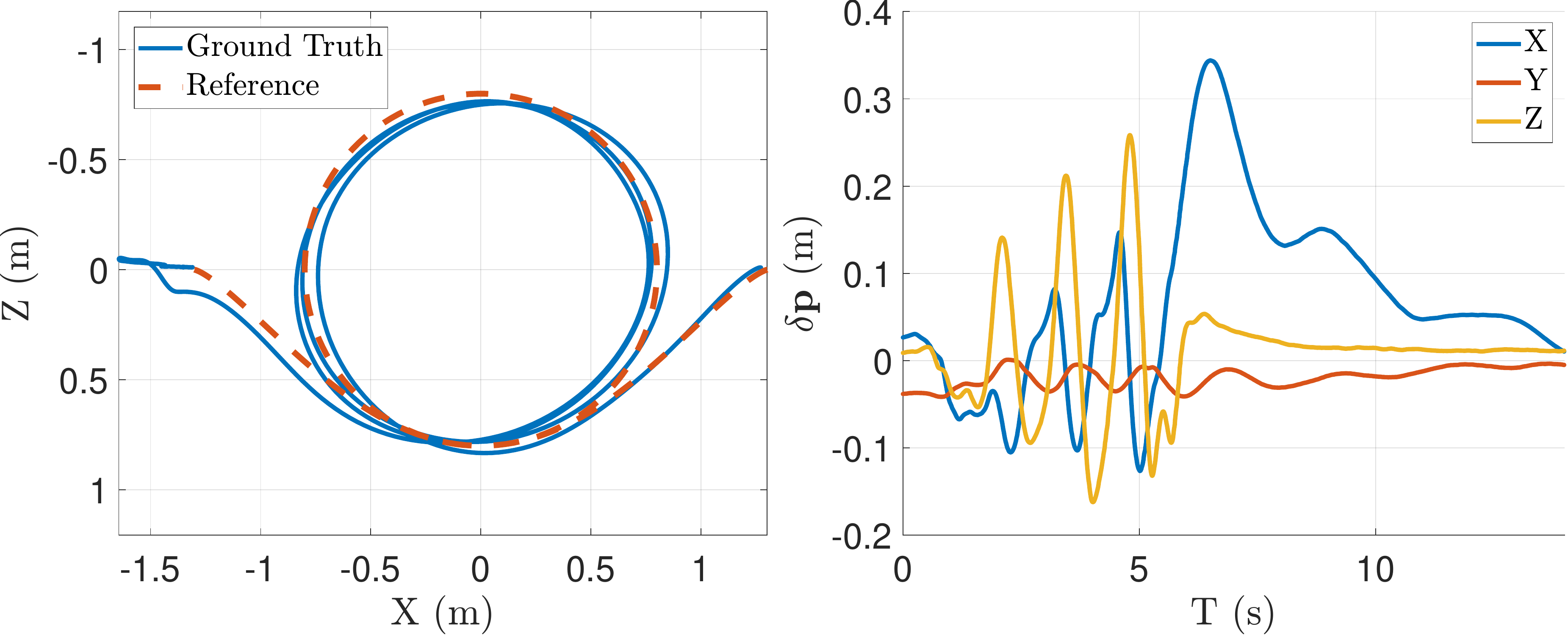}  
		\caption{Control performance of the quadrotor flies an acrobatic with triple vertical flipping.}  
		\label{fig:quad_virtical_circle}
		\setlength{\abovecaptionskip}{-0.1cm} 
        \setlength{\belowcaptionskip}{-0.2cm}
	\end{figure}

     As shown in Fig. \ref{fig:quad_horizontal_circle}, the quadrotor tracks the horizontal circle trajectory with radius of $1.3 m$ and speed increasing from $0 m/s$ to $5 m/s$, and acceleration peaking at $2g$. The position tracking error remains below $0.1 m$ even in such aggressive maneuver, demonstrating a significantly superior accuracy than that in the state-of-the-art MPC and LQR studies \cite{kamel2017linear}, \cite{bicego2019nonlinear}, \cite{foehn2018onboard}. The computation time for each MPC loop with complete process as shown in Algorithm \ref{a:mpc} is $5 ms$ in average, satisfying the real-time requirement. Fig. \ref{fig:quad_virtical_circle} shows the high tracking performance in a highly aggressive quadrotor acrobatic triple vertical flipping by the tracking trajectory and position error. We refer readers to the accompanying video at \url{https://youtu.be/wikgrQbE6Cs} for better demonstration of the two flights. 


	
    \subsection{Application to UGVs moving on Curved Surfaces}\label{sec:groundvehicle}
    
    
    \subsubsection{\textbf{\textit{Experimental Setup}}}
    We implement the proposed MPC framework on the chassis of a DJI RoboMaster ground vehicle (see Fig. \ref{fig:car_platform}), equipped with a Livox Avia LiDAR and a NUC10i7FNK (Intel i7, 1.1-4.7 GHz, 6 cores, 16GB RAM) onboard computer. The Mecanum wheels are equipped with rubber tires such that the vehicle is no lateral velocity. The robot provides longitudinal speed and steering (i.e., yaw) rate interfaces, which are the control input computed by the MPC controller. We use LiDAR-inertial Odometry FAST-LIO \cite{xu2021fast} running onboard the NUC computer to estimate the robot's state online. Both the FAST-LIO \cite{xu2021fast} and MPC run on the same onboard computer at $50Hz$. The MPC horizon is set to 45. We conduct the experiment on a road with significant height variation (see Fig. \ref{fig:car_mpc_performace}).

    
    \begin{figure}[!t] 
    	\centering 
    	\includegraphics[width=0.7\columnwidth]{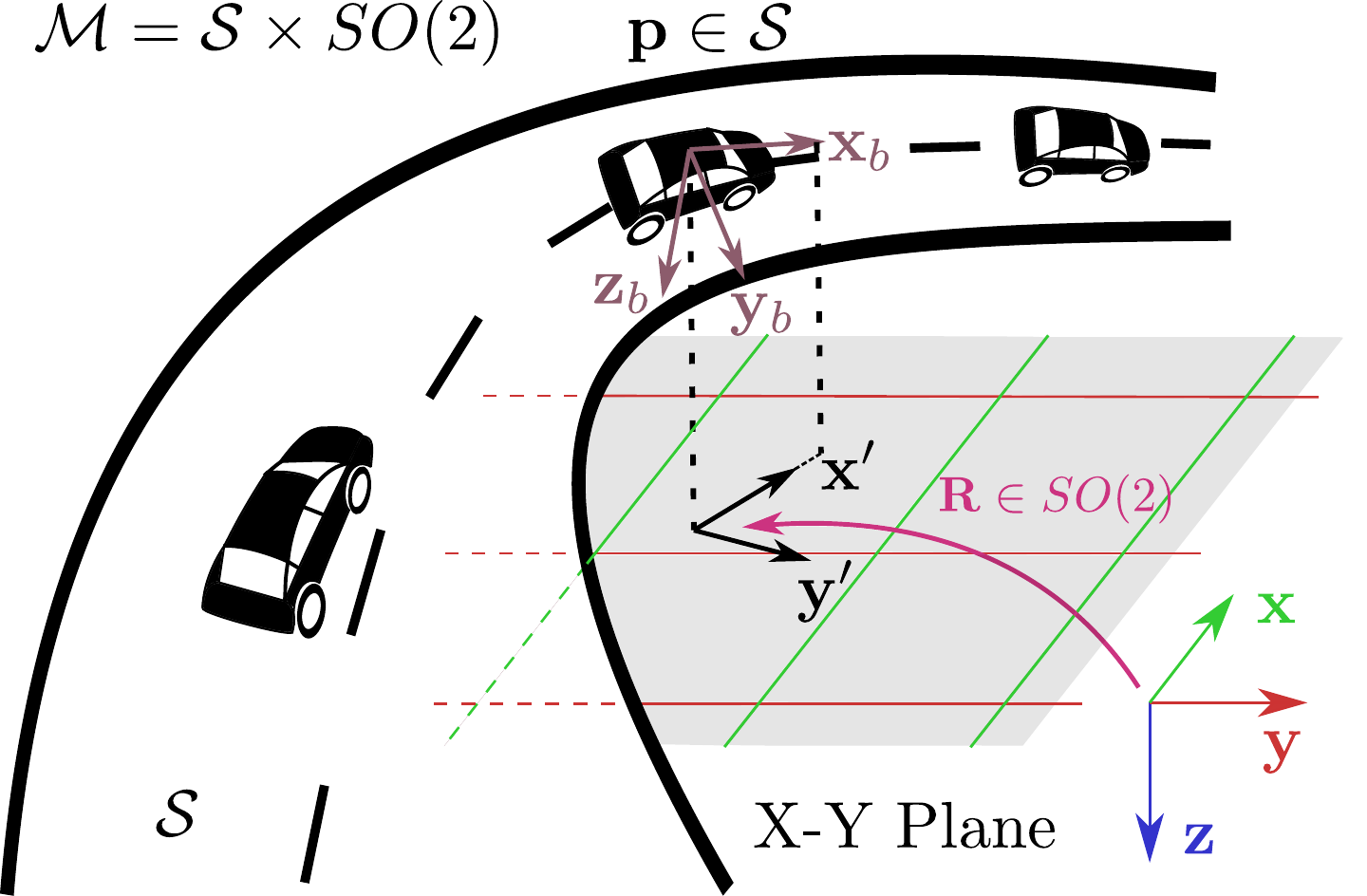}
    	\caption{Illustration of compound manifold $S \times SO(2)$.}
        \label{fig:S_SO(2)} 
        \setlength{\abovecaptionskip}{-0.1cm}
        \setlength{\belowcaptionskip}{-0.2cm}
    	\vspace{-0.1in}
    \end{figure}

    \subsubsection{\textbf{\textit{System Modeling}}}

    Shown in Fig. \ref{fig:S_SO(2)}, the UGV state is described by the position $\mathbf{p} \in \mathbb{R}^3$ in the fixed inertial frame denoted by $\mathbf x, \mathbf y, \mathbf z$ and its attitude represented by the body frame $\mathbf x_b, \mathbf y_b$ and $\mathbf z_b$. Note that since the UGV is restricted on the surface with the $\mathbf z_b$ always pointing to the surface normal, there is a one to one map between the body frame and the frame $\mathbf x', \mathbf y', \mathbf z$, which undergoes only a rotation around $\mathbf z$ from the inertial frame and can hence be represented by a rotation $\mathbf R \in SO(2)$. As a consequence, the robot is $\mathbf x = (\mathbf{p}, \mathbf R) \in \mathcal{S} \times SO(2)$. The control action of the robot is the velocity ${v}^\mathcal{B}_x$ along the body axis $\mathbf{x}_b$ and the yaw rate $\omega_z^{\mathcal{B}}$ along the body axis $\mathbf{z}_b$.
    
    Denote $\mathbf{x}_b( \mathbf{p}, \mathbf{R})$ the map from the frame $\mathbf x', \mathbf y', \mathbf z$ represented by $\mathbf R \in SO(2)$ to the body axis $\mathbf x_b$ and $\mathbf{z}_b(\mathbf{p})$ the body axis $\mathbf z_b$ which always points to the surface normal and hence depends on the robot's location on the surface. Then, the robot's model in continuous time is 
    
    \begin{subequations}
        \small
    	\begin{align}
	    	&\dot{\mathbf{p}} =  \mathbf{x}_b( \mathbf{p}, \mathbf{R}) {v}^{\mathcal{B}}_x \label{eq:vel}\\
	    	&\dot{\mathbf{R}} = \mathbf{R}\lfloor \mathbf{z}^T  \mathbf{z}_b(\mathbf{p})\omega_z^{\mathcal{B}} \rfloor  
    	\end{align}
    	\label{e:ugv_model_cont}
    	\setlength{\abovedisplayskip}{0.15cm} 
        \setlength{\belowdisplayskip}{0.15cm}
    \end{subequations}
    where we project the angular velocity $\omega_z^{\mathcal{B}}\mathbf{z}_b$ to the axis $\mathbf z$ to drive $\mathbf R$.

    \subsubsection{\textbf{\textit{Canonical Representation}}}
    
    We discretize the system (\ref{e:ugv_model_cont}) using zero order hold and represent it to the canonical form using the $\oplus$ defined respectively on $\mathcal{S}$ and $SO(2)$ in Section \ref{sec:operations}. For the translation, the velocity $\mathbf{x}_b( \mathbf{p}, \mathbf{R}) {v}^{\mathcal{B}}_x$ in (\ref{eq:vel}) causes a perturbation $\mathbf{x}_b( \mathbf{p}_k, \mathbf{R}_k) {v}^{\mathcal{B}}_{x_k} \Delta t \in \mathbb{R}^3$ over one sampling period, this perturbation should be projected to the $x-y$ plane that is described by $\oplus$. Hence, the translation equation is $\mathbf{p}_{k+1} = \mathbf{p}_{k} \oplus_{\mathcal{S}}(\mathbf E_{12}^T \mathbf{x}_b( \mathbf{p}_k, \mathbf{R}_k) {v}^{\mathcal{B}}_{x_k} \Delta t)$. Similarly for the rotation, the state equation is $\mathbf R_{k+1} =\mathbf R_{k} \oplus_{SO(2)} (\mathbf{z}^T  \mathbf{z}_b(\mathbf{p}_k)\omega_{z_k}^{\mathcal{B}} \Delta t)$. As a consequence, the UGV system in (\ref{e:ugv_model_cont}) can be discretized and cast into the canonical form as follows:
    \begin{subequations}
        \small
        \begin{align}
            &\mathcal{M} = \mathcal{S} \times SO(2), \quad \dim(\mathcal{M}) = 3 \\
            &\mathbf{x} = \begin{pmatrix}
                \mathbf{p} & \mathbf{R}
            \end{pmatrix} \in \mathcal{M}, \quad \mathbf{u} = \begin{bmatrix}
            {v}^\mathcal{B}_x & \omega^{\mathcal{B}}_z
            \end{bmatrix}  \in \mathbb{R}^2 \\
            &\mathbf{f}(\mathbf{x}, \mathbf{u}) \! = \! \begin{bmatrix}
                \mathbf E_{12}^T \mathbf{x}_b( \mathbf{p}, \mathbf{R}) {v}^{\mathcal{B}}_{x} \\
                \mathbf{z}^T  \mathbf{z}_b(\mathbf{p})\omega_{z}^{\mathcal{B}}
            \end{bmatrix} \! = \! \begin{bmatrix}
                \alpha \mathbf{R} \mathbbm{e}_1 {v}^{\mathcal{B}}_{x} \\
                \beta \omega^{\mathcal{B}}_{z}
            \end{bmatrix} \in \mathbb{R}^3
        \end{align}
        \setlength{\abovedisplayskip}{0.15cm} 
        \setlength{\belowdisplayskip}{0.15cm}
    \end{subequations}
    where 
    \begin{subequations}
        \small
    	\begin{align}
        &\alpha(\mathbf{p}, \mathbf{R}) = \frac{1}{\sqrt{1 + (\mathbf g^T \mathbf R\mathbbm{e}_1)^2 } }, \beta(\mathbf{p}) =\frac{1}{\sqrt{1 + \mathbf g^T \mathbf g}} \\
        &\mathbf g = \begin{bmatrix}
			 	F'_x & F'_y 
			\end{bmatrix}^T \in \mathbb{R}^{2}, \mathbbm{e}_1 = \begin{bmatrix} 1 & 0 \end{bmatrix}^T.
    	\end{align}
    	\setlength{\abovedisplayskip}{0.15cm} 
        \setlength{\belowdisplayskip}{0.15cm}
    \end{subequations}

     The manifold-specific parts of the system matrices are referred to TABLE \ref{tab_important_manifolds}, and the system-specific parts are given as follows:
    \begin{subequations}
        \footnotesize
        \begin{align}
            &\frac{\partial \mathbf{f}(\mathbf{x} \! \boxplus \! \delta\mathbf{x} , \mathbf{u} )}{\partial \delta\mathbf{x}} \! =\!  \begin{bmatrix}
                 \mathbf{R} \mathbbm{e}_1 {v}^{\mathcal{B}}_{x}\frac{\partial \alpha}{\partial \delta\mathbf{p}}  & \mathbf{R} \!\left(\frac{\partial \alpha}{\partial \delta\mathbf{R}}\mathbbm{e}_1 \! + \! \alpha \mathbbm{e}_2\right){v}^{\mathcal{B}}_{x} \\ \frac{\partial \beta}{\partial \delta\mathbf{p}} \omega^{\mathcal{B}}_{z} & 0
            \end{bmatrix} \in \mathbb{R}^{3 \times 3} \\
            &\frac{\partial \mathbf{f}(\mathbf{x}, \mathbf{u}\boxplus \delta\mathbf{u})}{\partial \delta\mathbf{u}} =  \begin{bmatrix}
                \alpha \mathbf{R} \mathbbm{e}_1 & \mathbf{0}_{2 \times 1} \\
                {0} & \beta
            \end{bmatrix} \in \mathbb{R}^{3 \times 2}
        \end{align}
        \setlength{\abovedisplayskip}{0.15cm} 
        \setlength{\belowdisplayskip}{0.15cm}
    \end{subequations}
	where
	\begin{subequations}
	    \footnotesize
		\begin{align}
			&\frac{\partial \alpha}{\partial \delta\mathbf{p}} = -\frac{\mathbf{g}^T \mathbf{R} \mathbbm{e}_1 \mathbbm{e}_1^T \mathbf R^T}{\sqrt{\left(1 + (\mathbf g^T \mathbf R\mathbbm{e}_1)^2 \right)^3} } \frac{\partial \mathbf{g}\left(\mathbf{p} \boxplus \delta \mathbf{p} \right)}{\partial \delta \mathbf{p}} \in \mathbb{R}^{1 \times 2} \\
			&\frac{\partial \alpha}{\partial \delta\mathbf{R}} = -\frac{ \mathbbm{e}_1^T \mathbf{R}^T \mathbf{g} \mathbf{g}^T}{\sqrt{\left(1 + (\mathbf g^T \mathbf R\mathbbm{e}_1)^2 \right)^3} } \frac{\partial \left( (\mathbf R \boxplus \delta \mathbf R) \mathbbm{e}_1 \right)}{\partial \delta \mathbf R} \in \mathbb{R} \\
			&\frac{\partial \beta}{\partial \delta\mathbf{p}} = -\frac{\mathbf g^T}{\sqrt{\left(1 + \mathbf g^T \mathbf g\right)^3}}  \frac{\partial \mathbf{g}\left(\mathbf{p} \boxplus \delta \mathbf{p} \right)}{\partial \delta \mathbf{p}} \in \mathbb{R}^{1 \times 2} \\
			&\frac{\partial \mathbf{g}\left(\mathbf{p} \boxplus \delta \mathbf{p} \right)}{\partial \delta \mathbf{p}} \! = \!  \begin{bmatrix} F''_{xx} & F''_{xy} \\ F''_{yx} & F''_{yy} \end{bmatrix}, \frac{\partial \left( (\mathbf R \! \boxplus \!  \delta \mathbf R) \mathbbm{e}_1 \right)}{\partial \delta \mathbf R} \!  =  \!  \mathbf R \mathbbm{e}_2.
		\end{align}
		\setlength{\abovedisplayskip}{0.15cm} 
        \setlength{\belowdisplayskip}{0.15cm}
	\end{subequations}


     \subsubsection{\textbf{\textit{Trajectory generation and Surface Approximation}}}
    To generate a reference trajectory on the ground surface, we first manually control the ground vehicle to travel along the road and run FAST-LIO to build a 3D point cloud map of the traveled environment (Fig. \ref{f:quad_car_traj}). Then a 2D smooth trajectory on the $X$-$Y$ plane is generated and sampled. At each sampled point, the orientation trajectory (i.e., $\mathbf R^d \in SO(2)$) is determined as being tangent to the 2D path. Two consecutive orientations determine the raw rate, which are mapped to the expected UGV body yaw rate $\omega_{z_d}^{\mathcal{B}}$ by dividing by $\mathbf z^T \mathbf z_b(\mathbf p^d)$. Moreover, for each sampled 2D point, we search the point cloud  map for neighbouring points whose $(x,y)$ coordinates are nearby. The center of these neighbouring points then defines a trajectory point on the ground surface (i.e., $\mathbf p^d \in \mathcal{S}$). The neighbouring points are then used to approximate a smooth surface around the center by a quadratic polynomial \cite{zhang2021pose}:
    \begin{eqnarray}
        F(x,y) = \gamma_1 x^2 + \gamma_2 xy + \gamma_3 y^2 + \gamma_4 x + \gamma_5 y + \gamma_6
    \end{eqnarray}
    where the surface parameter $\boldsymbol{\gamma} = [\gamma_1 \  \gamma_2 \ \gamma_3 \  \gamma_4 \  \gamma_5 \ \gamma_6]^T$ is estimated by least square method. The fitted surface allows to evaluate the value of $F_x', F_y', F''_{xx}$ and $F''_{yy}$ as required by the system specific parts. In this experiment, we expect the UGV speed $v_{x_d}^{\mathcal{B}}$ on the ground surface to be constant. Hence, we sample the next point on the 2D trajectory such that the corresponding 3D trajectory point on the surface is $v_{x_d}^{\mathcal{B}}\Delta t$ away from the current one. 
    
    \begin{figure}[!t]
        \centering
        \includegraphics[width=1\columnwidth]{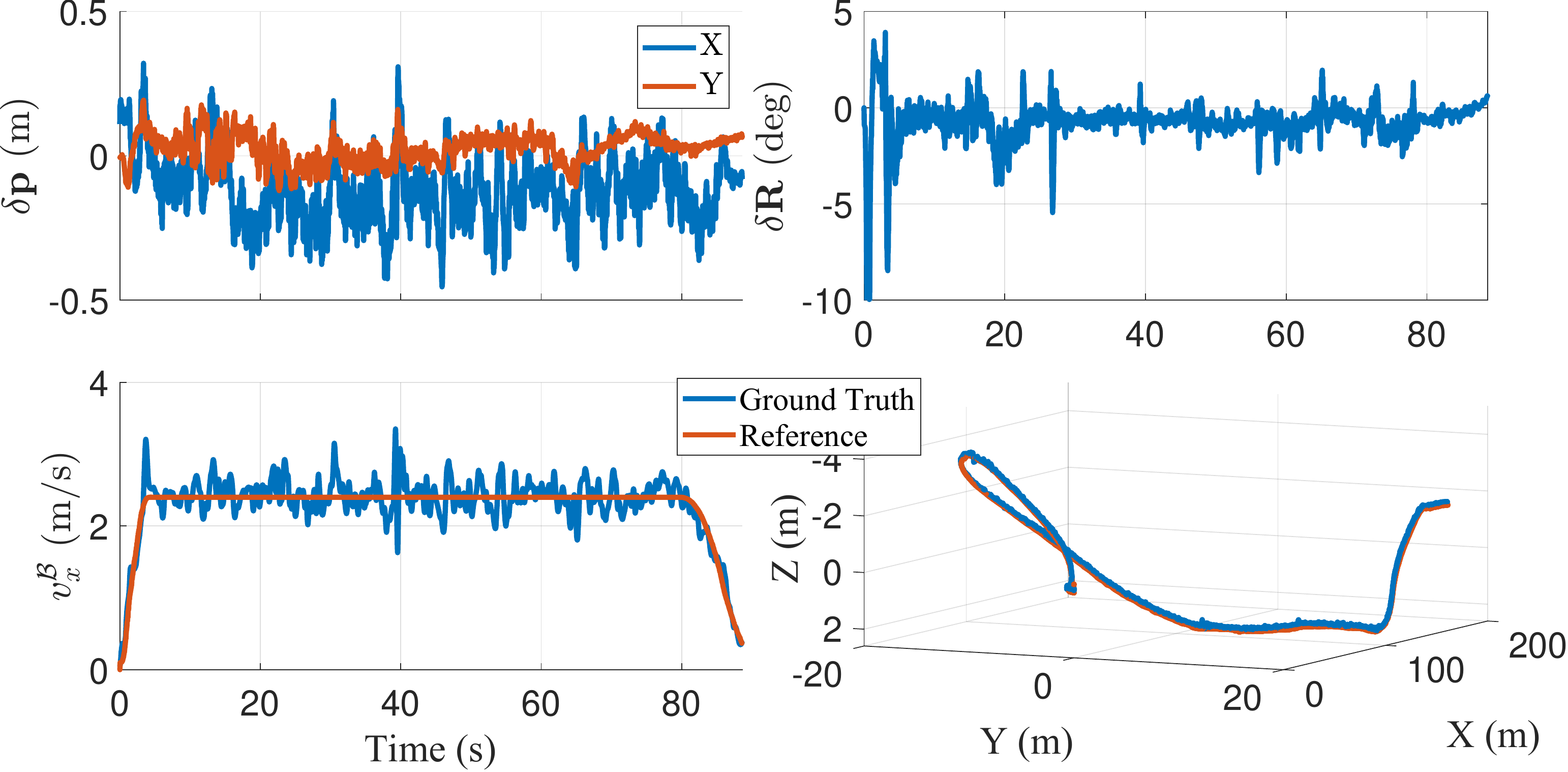}
        \caption{Control performancre of the UGV moving on curved surface with a constant speed of $2.4 m/s$.}
        \label{fig:car_mpc_performace}
        \setlength{\abovecaptionskip}{-0.1cm}
        \setlength{\belowcaptionskip}{-0.2cm}
    	\vspace{-0.1in}
    \end{figure}

    \subsubsection{\textbf{\textit{Experimental Results}}}
    As shown in Fig. \ref{f:quad_car_traj} and the supplemental video, firstly, a  reference trajectory that is smooth and well-fit the surface is generated in real time, resulting from the precious lidar point cloud map and surface approximation. Showing in Fig. \ref{fig:car_mpc_performace}, the control error including position and orientation error are shown slight that the UGV tracks the reference position and speed of $2.4 m/s$ with sufficient accuracy, even on the road with significant height variation and sharp turn.

	\section{Conclusion and Future Work}
	\label{sec:conclusion}
 	This paper proposes a generic MPC formulation for trajectory tracking on manifolds, bridging the gap of  geometric tracking control and MPC implementation for robotic systems. A canonical, symbolic, minimal-parametrization, singularity-free framework with modeling, error-system derivation, linearization and control is established. The proposed algorithm is implemented on quadrotor UAV with aggressive maneuvers and UGV moving on curved surfaces, showing a high tracking accuracy and computational efficiency. One potential challenge for this scheme is that the precision of linearizing the error system along the reference trajectory deteriorates when the initial tracking error is large. One of the solution is to shift the linearization along the actual trajectory but increase the computation cost due to the nonlinear complexity, while another method is to replan the reference trajectory if the tracking error is too large. The future work will investigate this challenge and its potential solutions.

 	\appendix
    \label{appendix}
 	We detail the derivation of the manifold-specific parts $\mathbf{G}_{\mathbf{x}}, \mathbf{G}_{\mathbf{f}}$ defined in (\ref{Gx_Gf}) of the error-state system matrices on the primitive manifolds. Denote $\mathbf{v} =  \Delta t \mathbf{f}(\mathbf{x}_k^d, \mathbf{u}_k^d)$ for simplicity.
 	
 	\subsection{Example 1: Euclidean Space $\mathbb{R}^n$}
 	\begin{subequations}
 	    \small
 		\begin{align}
 			\mathbf{G}_{\mathbf{x}_k} &= \frac{\partial \left(\mathbf{x}_k^d + \boldsymbol{\delta} + \mathbf{v} - \left(\mathbf{x}_k^d + \mathbf{v}\right)\right) }{\partial \boldsymbol{\delta}} = \mathbf{I}_n \\
 			\mathbf{G}_{\mathbf{f}_k} &= \frac{\partial\left(\mathbf{x}_k^d + \mathbf{v} + \boldsymbol{\delta}- \left(\mathbf{x}_k^d + \mathbf{v}\right)\right) }{\partial\boldsymbol{\delta}} = \mathbf{I}_n
 		\end{align}
 		\setlength{\abovedisplayskip}{0.15cm} 
        \setlength{\belowdisplayskip}{0.15cm}
 	\end{subequations}
 
 	\subsection{Example 2.1: Special Orthogonal Group $SO(2)$}
	
	\begin{subequations}
		\small
		\begin{align}
		    \mathbf{G}_{\mathbf{x}_k} &= \frac{\partial \text{Log}\left( \left(\mathbf{x}_k^d \cdot \text{Exp}(v) \right)^{-1} \mathbf{x}_k^d \cdot \text{Exp}(\delta) \cdot \text{Exp}(v) \right)  }{\partial \delta} \nonumber\\
    		&\approx \frac{\partial  \text{Log} \left(\text{Exp}(\delta) \right)}{\partial \delta} = 1 \\
    		\mathbf{G}_{\mathbf{f}_k} &\approx \frac{\partial  \text{Log} \left( \left(\mathbf{x}_k^d \text{Exp}(v) \right)^{-1} \mathbf{x}_k^d \text{Exp}(v + \delta) \right)}{\partial \delta}  \nonumber\\
    		&\approx \frac{\partial  \text{Log} \left(\text{Exp}(\delta) \right)}{\partial \delta} = 1
		\end{align}
		\setlength{\abovedisplayskip}{0.15cm} 
        \setlength{\belowdisplayskip}{0.15cm}
	\end{subequations}
	
	\subsection{Example 2.2: Special Orthogonal Group $SO(3)$}
 	\begin{subequations}
 		\small
 		\begin{align}
 		\mathbf{G}_{\mathbf{x}_k} &= \frac{\partial}{\partial \boldsymbol{\delta}} \text{Log}\left( \left(\mathbf{x}_k^d \cdot \text{Exp}(\mathbf{v}) \right)^{-1} \mathbf{x}_k^d \cdot \text{Exp}(\boldsymbol{\delta}) \cdot \text{Exp}(\mathbf{v}) \right) \nonumber\\
 		&\approx  \frac{\partial }{\partial \boldsymbol{\delta}} \text{Log} \left(\text{Exp}\left( \text{Exp}(\mathbf{v})^{-1} \boldsymbol{\delta}  \right) \right) = \frac{\partial }{\partial \boldsymbol{\delta}} \left( \text{Exp}(\mathbf{v})^{-1} \boldsymbol{\delta}  \right)  \nonumber\\
 		&= \text{Exp}(-\mathbf{v}) \\
 		\mathbf{G}_{\mathbf{f}_k} &=\frac{\partial}{\partial \boldsymbol{\delta}} \text{Log}\left( \left(\mathbf{x}_k^d \cdot \text{Exp}(\mathbf{v}) \right)^{-1} \mathbf{x}_k^d \cdot \text{Exp}(\mathbf{v} + \boldsymbol{\delta}) \right) \nonumber\\
 		&=\frac{\partial}{\partial \boldsymbol{\delta}} \text{Log}\left( \text{Exp}(-\mathbf{v}) \cdot \text{Exp}(\mathbf{v} + \boldsymbol{\delta}) \right)  = \frac{\partial }{\partial \boldsymbol{\delta}} \left( \mathbf{A}(\mathbf{v})^T \boldsymbol{\delta}  \right) \nonumber \\
 		&= \mathbf{A}(\mathbf{v})^T
 		\end{align}
 		\setlength{\abovedisplayskip}{0.15cm} 
        \setlength{\belowdisplayskip}{0.15cm}
 	\end{subequations}
 	where 
	\begin{equation}
		\small
		\mathbf{A}(\mathbf{\boldsymbol{\theta}}) = \mathbf{I}_3 + \left(\frac{1-\cos\Vert\boldsymbol{\theta}\Vert}{\Vert\boldsymbol{\theta}\Vert}\right) \frac{\lfloor\boldsymbol{\theta}\rfloor}{\Vert\boldsymbol{\theta}\Vert} + \left(1 - \frac{\sin\Vert\boldsymbol{\theta}\Vert}{\Vert\boldsymbol{\theta}\Vert} \right) \frac{\lfloor\boldsymbol{\theta}\rfloor^2}{\Vert\boldsymbol{\theta}\Vert^2}
		\setlength{\abovedisplayskip}{0.15cm} 
        \setlength{\belowdisplayskip}{0.15cm}
	\end{equation}
 
 	\subsection{Example 3: Two-Sphere $\mathbb{S}^2_r$}
     Referring to our previous work \cite{he2021embedding}, we have
    \begin{subequations}
        \small
        \begin{align}
            \mathbf{G}_{\mathbf{x}_k} &= \left(\frac{\partial \mathbf{x} \boxminus \mathbf{y}}{\partial \mathbf{x}} \cdot \frac{\partial \mathbf{x}}{\partial \boldsymbol{\delta}}\right) \bigg|_{\begin{subarray}{l}\mathbf{x} = \text{Exp}(\mathbf{v})\text{Exp}(\mathbf{B}(\mathbf{x}_k^d)\boldsymbol{\delta})\cdot\mathbf{x}_k^d \\ \mathbf{y} = \mathbf{x}_{k}^d \oplus \mathbf{v}, \boldsymbol{\delta} = \mathbf{0}\end{subarray}} \nonumber\\
            &\approx \frac{1}{r^2}\mathbf{B}(\mathbf{x}_{k}^d \oplus \mathbf{v})^T \text{Exp}(\mathbf{v})\lfloor \mathbf{x}_{k}^d \rfloor \text{Exp}(\mathbf{v})^T \nonumber\\
            &\qquad \cdot \frac{\partial}{\partial \boldsymbol{\delta} }\left(\text{Exp}(\mathbf{v}) \left( \mathbf{I}_3 + \lfloor \mathbf{B}(\mathbf{x}_k^d) \boldsymbol{\delta} \rfloor \right)  \cdot\mathbf{x}_k^d\right) \nonumber\\
            &= -\frac{1}{r^2}\mathbf{B}(\mathbf{x}_{k}^d \oplus \mathbf{v})^T \text{Exp}(\mathbf{v})\lfloor \mathbf{x}_{k}^d \rfloor^2 \mathbf{B}(\mathbf{x}_k^d) \\
            \mathbf{G}_{\mathbf{f}_k} &= \left(\frac{\partial \mathbf{x} \boxminus \mathbf{y}}{\partial \mathbf{x}} \cdot \frac{\partial \mathbf{x}}{\partial \boldsymbol{\delta} }\right) \bigg|_{\begin{subarray}{l}\mathbf{x} = \text{Exp}(\mathbf{v}+\boldsymbol{\delta})\cdot\mathbf{x}_k^d \\ \mathbf{y} = \mathbf{x}_{k}^d \oplus \mathbf{v}, \boldsymbol{\delta} = \mathbf{0}\end{subarray}} \nonumber \\
            &= -\frac{1}{r^2}\mathbf{B}(\mathbf{x}_{k}^d \oplus \mathbf{v})^T \text{Exp}(\mathbf{v})\lfloor \mathbf{x}_{k}^d \rfloor^2 \mathbf{A}(\mathbf{v})^T   \end{align}
        \setlength{\abovedisplayskip}{0.15cm} 
        \setlength{\belowdisplayskip}{0.15cm}
    \end{subequations}
 	
 	\subsection{Example 4: 2D Surface $\mathcal{S}$}
 	\begin{equation}
 		\small
 		\begin{aligned}
	 		\mathbf{G}_{\mathbf{x}_k} &= \frac{\partial }{\partial \boldsymbol{\delta}} \mathbf{E}_{12}^T \left( \left(\mathbf{x}_{k}^d \boxplus \boldsymbol{\delta}    \right) \oplus \mathbf{v} - \left(\mathbf{x}_{k}^d \oplus \mathbf{v}\right)\right) \\
	 		&= \frac{\partial }{\partial \boldsymbol{\delta}} \mathbf{E}_{12}^T \left( \begin{bmatrix} 
	 		\mathbf{E}_{12}^T \left(\mathbf{x}_{k}^d \boxplus \boldsymbol{\delta} \right)   + \mathbf{v} \\
	 		F\left( \mathbf{E}_{12}^T \left(\mathbf{x}_{k}^d \boxplus \boldsymbol{\delta} \right)   + \mathbf{v} \right)
	 		\end{bmatrix}- \begin{bmatrix} 
	 		\mathbf{E}_{12}^T\mathbf{x}_{k}^d    + \mathbf{v} \\
	 		F\left( \mathbf{E}_{12}^T\mathbf{x}_{k}^d    + \mathbf{v} \right)
	 		\end{bmatrix}\right) \\
	 		&= \frac{\partial }{\partial \boldsymbol{\delta}} \left(\mathbf{E}_{12}^T \left(\mathbf{x}_{k}^d \boxplus \boldsymbol{\delta} \right) + \mathbf{v} - \mathbf{E}_{12}^T\mathbf{x}_{k}^d - \mathbf{v} \right) \\
	 		&= \frac{\partial }{\partial \boldsymbol{\delta}} \left(\mathbf{E}_{12}^T \begin{bmatrix} 
	 		\mathbf{E}_{12}^T\mathbf{x}_{k}^d    + \boldsymbol{\delta} \\
	 		F\left( \mathbf{E}_{12}^T\mathbf{x}_{k}^d  + \boldsymbol{\delta} \right)
	 		\end{bmatrix} - \mathbf{E}_{12}^T\mathbf{x}_{k}^d \right) \nonumber\\
	 		&= \frac{\partial }{\partial \boldsymbol{\delta}} \left(\mathbf{E}_{12}^T\mathbf{x}_{k}^d    + \boldsymbol{\delta} - \mathbf{E}_{12}^T\mathbf{x}_{k}^d \right) = \mathbf{I}_2
 		\end{aligned}
 		\setlength{\abovedisplayskip}{0.15cm} 
        \setlength{\belowdisplayskip}{0.15cm}
 	\end{equation}
 	\begin{equation}
 		\small
 		\begin{aligned}
	 		\mathbf{G}_{\mathbf{f}_k} &= \frac{\partial }{\partial \boldsymbol{\delta}} \mathbf{E}_{12}^T \left( \mathbf{x}_{k}^d   \oplus \left( \mathbf{v} + \boldsymbol{\delta} \right) - \left(\mathbf{x}_{k}^d \oplus \mathbf{v}\right)\right) \\
	 		&= \frac{\partial }{\partial \boldsymbol{\delta}} \mathbf{E}_{12}^T \left( \begin{bmatrix} 
	 		\mathbf{E}_{12}^T\mathbf{x}_{k}^d    + \mathbf v + \boldsymbol{\delta} \\
	 		F\left( \mathbf{E}_{12}^T\mathbf{x}_{k}^d    + \mathbf v + \boldsymbol{\delta}  \right)
	 		\end{bmatrix}  - \begin{bmatrix} 
	 		\mathbf{E}_{12}^T\mathbf{x}_{k}^d    + \mathbf v  \\
	 		F\left( \mathbf{E}_{12}^T\mathbf{x}_{k}^d    + \mathbf v  \right)
	 		\end{bmatrix} \right) \\
	 		&= \frac{\partial }{\partial \boldsymbol{\delta}}  \left( \mathbf{E}_{12}^T\mathbf{x}_{k}^d    + \mathbf v + \boldsymbol{\delta}  - \mathbf{E}_{12}^T\mathbf{x}_{k}^d    - \mathbf v \right) = \mathbf{I}_2
 		\end{aligned}
 		\setlength{\abovedisplayskip}{0.15cm} 
        \setlength{\belowdisplayskip}{0.15cm}
 	\end{equation}

    \bibliographystyle{IEEEtran}
    \bibliography{reference}

\end{document}